\documentclass[10pt]{article}
\usepackage[preprint]{tmlr}

\usepackage{amsmath,amsfonts,bm}

\usepackage{hyperref}
\usepackage{url}

\usepackage{amsthm}
\usepackage{algorithm}

\usepackage{algorithmic}

\usepackage{subfiles}
\usepackage{graphicx}
\usepackage{enumitem}
\usepackage{seqsplit}
\usepackage{booktabs}
\usepackage{xcolor}

\title{PLDiv: Measuring Dataset Diversity and Semantic Collapse via Persistence Landscapes}

\author{%
\name Yang Ba \email yangba@asu.edu \\
\addr School of Computing and Augmented Intelligence\\
Arizona State University
\\[12pt]
\name Michelle V. Mancenido \email mmanceni@asu.edu \\
\addr School of Mathematical and Natural Sciences \\
Arizona State University
\\[12pt]
\name Rong Pan \email Rong.Pan@asu.edu\\
\addr School of Computing and Augmented Intelligence\\
Arizona State University
}

\begin{document}

\maketitle

\begin{abstract}
Modern data pipelines increasingly combine human-created and human-curated data with AI-generated data for model training and evaluation. As datasets grow, they often add semantically redundant samples rather than expanding semantic coverage, which deteriorate the quality of the whole dataset. Existing diversity metrics primarily summarize lexical variation, entropy, or pairwise similarity, but provide limited characterization of the shape of the underlying data distribution. In this paper, we propose PLDiv, a diversity metric derived from persistence landscapes that quantifies the structural organization of a dataset from its geometric representation. PLDiv summarizes how data neighborhoods evolve as nearby samples progressively connect, producing an interpretable scalar measure of structural diversity without requiring task-specific supervision.
Whether measuring intrinsic diversity in text generations or image embeddings, or detecting semantic collapse in instruction datasets and reasoning trajectories, PLDiv performs competitively against evaluated baselines and aligns strongly with diversity labels during semantic collapse. Consequently, it effectively supports applications in synthetic data monitoring, dataset construction, data augmentation, and generative model evaluation. The code is available at https://github.com/YangBa78/PLDiv.

\end{abstract}

\section{Introduction}

Large language models (LLMs) are increasingly reliant on synthetic data generated by peer systems across the complete development lifecycle, including pretraining, alignment, benchmarking, and iterative deployment. This creates a data quality challenge: while models may maintain fluent and competent outputs, the diversity of responses, instructions, and ideas can steadily narrow. Such homogenization is evident in open-ended generation, where models tend to converge on similar phrasings, topics, and reasoning patterns despite many plausible alternatives \citep{padmakumar2023does,jiang2025artificial,gude-etal-2026-aligned}. This issue also affects instruction tuning and synthetic data creation \citep{wang2022self,zhou2023lima}, where large datasets may still suffer from limited semantic coverage due to many near-duplicate or semantically similar examples \citep{lee-etal-2022-deduplicating}.

Measuring diversity is therefore essential to ensure LLMs capture genuinely distinct intents and reasoning patterns. It reveals whether generated outputs reflect meaningful semantic variation or merely superficial rewording, and whether synthetic data pipelines collapse toward narrow modes through repeated sampling, filtering, or preference optimization. Diversity is not just a byproduct of decoding strategies; it reflects intrinsic properties of model reasoning processes \citep{xu-zhang-2026-language}. Moreover, recent work shows that homogenization can be self-reinforcing: as increasingly uniform generations enter training corpora, subsequent models further contract linguistic and conceptual variation \citep{bertrand2023stability,alemohammad2024self,shumailov2024ai}.

Existing diversity metrics capture important but limited aspects of this problem. Lexical measures like Distinct-$n$ \citep{li-etal-2016-diversity,song2024scaling} and Self-BLEU \citep{zhu2018texygen,shu2019generating} are easy to compute and widely used, but focus on token overlap and can miss semantic collapse when different words express the same idea. Embedding-based measures such as the Vendi Score \citep{dan2023vendi}, MagArea \citep{limbeck2024metric}, and DCScore \citep{zhu2025measuring} go beyond surface forms by analyzing similarity or distance matrices. However, these methods summarize entropy, scale, or separability without explicitly characterizing the shape of the underlying data distribution, limiting their sensitivity to distinctions between local clusters of near-duplicates and broader semantic coverage.

\begin{figure*}[!t]
    \centering
    \includegraphics[width=1\linewidth]{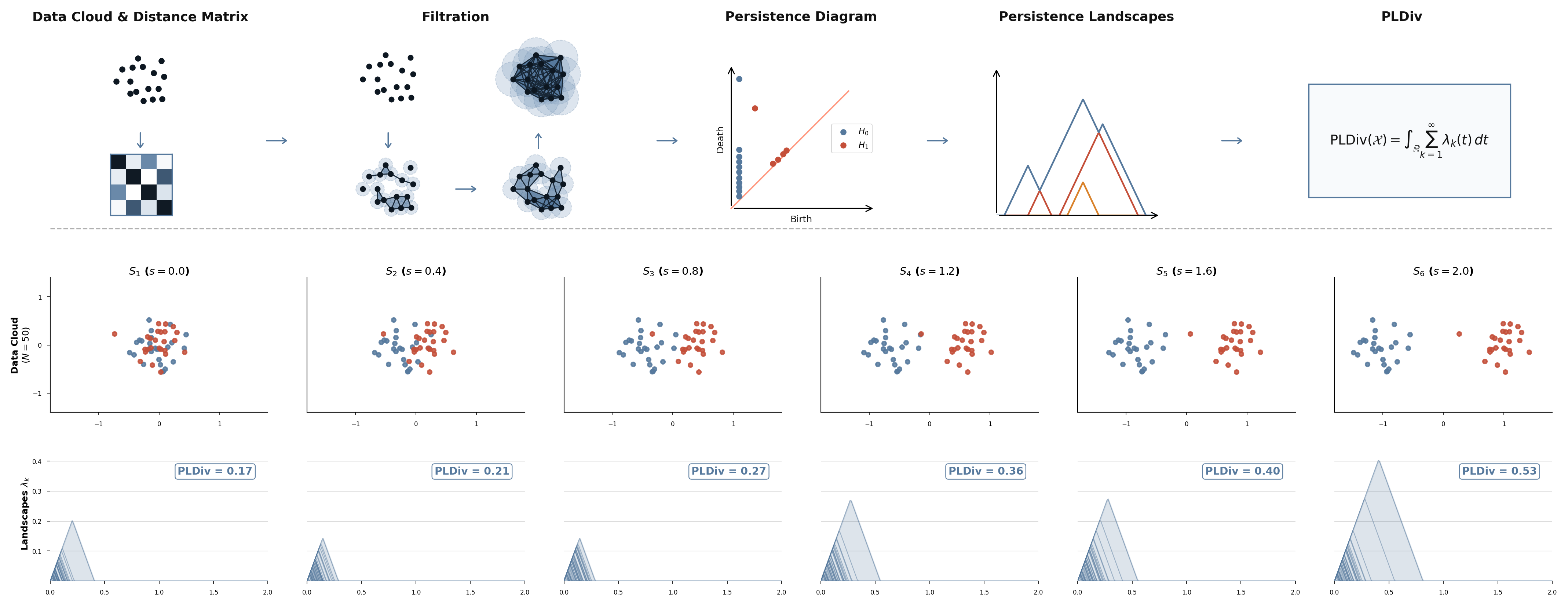}
\caption{
\textbf{Illustration of PLDiv conceptual pipeline (Top) and its behavior across synthetic Gaussian separation simulation states ($S_1$-$S_6$, Bottom).}Top: The topological workflow mapping an input data cloud/distance matrix $\mathbf{D}$ through filtration, persistence diagrams, and persistence landscapes to the scalar $\mathrm{PLDiv}$ diversity metric. Bottom: Synthetic evaluation demonstrating how PLDiv dynamically scales with cluster separation distance ($s \in [0.0, 2.0]$).}
\label{fig:pipeline}
\end{figure*}

We address this gap with a \textbf{persistence landscape-based diversity (PLDiv)} measure, a reference-free, geometry-aware diversity metric derived from topological data analysis (TDA). Given a dataset or its distance matrix, PLDiv integrates the total area of the persistence landscape and thereby summarizes the number and lifetimes of nonzero persistent features in a single score (Figure~\ref{fig:pipeline}). Its closed form assigns each feature a contribution equal to one quarter of its squared lifetime, emphasizing longer-lived structural organization while retaining the cumulative contribution of multiple features. The definition applies to different homology dimensions: $0$-dimensional PLDiv summarizes how separated components merge as the filtration grows, whereas $1$-dimensional PLDiv summarizes persistent loop-like structures. This allows PLDiv to characterize both dataset diversity and semantic collapse across different forms of structural variation. We further establish its diversity properties and develop a scalable sparse estimator for computing $0$-dimensional PLDiv on large datasets.

We evaluate PLDiv through experiments spanning complementary diversity scenarios. Gaussian separation and PCA reconstruction test its response to known geometric changes; text-generation benchmarks use decoding temperature and human judgments as diversity targets; a fixed-size instruction benchmark progressively introduces local semantic redundancy; a reasoning-trajectory experiment evaluates whether PLDiv captures semantic collapse induced by the homogenization of alternative solution paths; and a fixed-size image benchmark expands class coverage. Together, these experiments examine structural complexity, semantic coverage, and collapse across data representations, with comparisons against lexical, spectral, separability-based, and magnitude-based diversity measures.

Our contributions are threefold:
\begin{itemize}[topsep=0pt, partopsep=0pt, itemsep=1pt, parsep=0pt]
    \item We formulate PLDiv as a reference-free scalar summary of persistence landscapes. Its closed form gives an interpretable quadratic weighting of persistence lifetimes and applies to structural features in any chosen homology dimension.

    \item We characterize PLDiv axiomatically through effective size, monotonicity, twin invariance, and symmetry. We further develop and empirically validate a scalable sparse estimator for $0$-dimensional PLDiv on large embedding datasets.

    \item We empirically show that PLDiv consistently tracks the corresponding diversity targets in simulations, text and image embeddings, and instruction data and reasoning trajectory collapse, suggesting its potential utility for evaluating generative model outputs, synthetic data pipelines, dataset construction, and augmentation.

\end{itemize}

\section{Related Work}

\subsection{Diversity Measurement}

In the field of natural language processing (NLP) and natural language generation (NLG), diversity is commonly measured via lexical or pairwise variation, including Distinct-1/2 \citep{li-etal-2016-diversity}, distinct-$n$ \citep{song2024scaling}, Self-BLEU or inverse Self-BLEU \citep{zhu2018texygen,shu2019generating}, and ROUGE-L based analyses \citep{wang2022self,padmakumar2023does}.
Human evaluations show that these scores capture form-level variation better than semantic diversity \citep{tevet-berant-2021-evaluating}, motivating semantic metrics such as NLI-based diversity \citep{stasaski2022semantic} and dataset quality indicators such as DQI \citep{mishra2020dqi}. However, these token-based methods do not directly characterize meaning across embedding spaces.

Embedding-based diversity metrics go beyond surface overlap. Reference-based methods like FID \citep{heusel2017gans}, MAUVE \citep{pillutla2021mauve}, and precision–recall variants \citep{kynkaanniemi2019improved,bronnec2024exploring} compare generated data to gold corpora but often mix fidelity with diversity.
Reference-free metrics operate on embeddings or similarity matrices: diversity, density, and homogeneity \cite{lai2020diversity} summarize dispersion; Vendi Score \citep{dan2023vendi} and Renyi Kernel Entropy \citep{jalali2023information} use spectral entropy; DCScore \citep{zhu2025measuring} measures separability; and magnitude-based metrics estimate effective dataset size \citep{limbeck2024metric}. Classic cases like Anscombe's Quartet show that identical statistics can mask distinct geometric structures \citep{anscombe1973graphs,matejka2017same}. While useful, these methods often reduce structure to global summaries, limiting insight into local neighborhoods versus global coverage.

\subsection{Persistent Homology}

Persistent Homology (PH) \citep{edelsbrunner2002topological,edelsbrunner2008persistent} is a core tool in TDA for summarizing the shape of point-cloud data across scales. By building a filtration of simplicial complexes, PH records the birth and death of connected components, loops, and voids as barcodes or persistence diagrams, yielding a stable multi-scale description of data structure.

Beyond barcodes and persistence diagrams, persistence is connected to several established scalar and geometric summaries. Classical work characterized the asymptotic behavior of power-weighted Euclidean minimum-spanning trees \citep{steele1988growth}, which subsequently supported graph-based nonparametric estimation of R\'enyi entropy \citep{hero1999asymptotic}. Within persistent homology, \citet{cohensteiner2010lipschitz} studied degree-$p$ total persistence and established its stability under suitable regularity conditions. For the Vietoris--Rips $H_0$ construction used in this work, these two lines coincide: degree-$p$ total persistence reduces to a power-weighted minimum-spanning-tree functional. More recently, average persistence landscapes have been used to infer curvature from point samples on constant-curvature disks \citep{bubenik2020persistent}, while persistent magnitude provides a weighted, signed scalar invariant of persistence modules \citep{govc2021persistent}.

Persistent homology has also been applied directly to generative-model evaluation. Geometry Score compares the relative living times of $H_1$ features between reference and generated samples to detect mode collapse \citep{khrulkov2018geometry}, while TopP\&R uses topological and statistical significance to construct robust support estimates from which precision and recall are computed \citep{kim2023toppr}. Both methods compare generated samples against reference data. Building on these established connections, we formulate PLDiv as a reference-free dataset diversity measure and investigate its axiomatic properties, scalable estimation, and empirical behavior across semantic and structural diversity settings.

\section{Preliminaries}
\subsection{Persistence Diagrams}

PH provides a multiscale description of the topological structure of data.
Starting from a point cloud $\mathcal{X} = \{x_1, \dots, x_n\}$, it builds a nested sequence of simplicial complexes (a filtration), such as the Vietoris--Rips filtration.
This filtration can be understood intuitively by growing balls (or ``bubbles'') around each data point as $\epsilon$ increases. Under the Vietoris--Rips convention used in this work, $\epsilon$ represents the distance threshold between points, so the corresponding balls have radius $\epsilon/2$. When the balls around two points overlap, the two points are connected; as more balls overlap pairwise, higher-dimensional simplices are formed (see Figure~\ref{fig:pipeline}).
In this process, new topological features such as connected components, loops, and voids appear and eventually vanish when the bubbles merge or fill in.
This viewpoint highlights that persistent homology captures how the topology of the data evolves across scales of the underlying radius parameter.

Formally, each topological feature is associated with a birth time $b_i$, the smallest radius at which it appears, and a death time $d_i$, the radius at which it disappears (for instance, when two connected components merge or when a loop becomes filled).
The difference $\ell_i = d_i - b_i$ is called the \emph{lifetime} (or persistence) of the feature and quantifies its robustness across scales.

The output of persistent homology is summarized in a \emph{persistence diagram}, defined as the multiset
\[
\mathcal{D} = \{ (b_i, d_i) \}_{i=1}^m, \quad b_i < d_i,
\]
where each point $(b_i, d_i)$ represents the birth and death scales of a feature.
The diagram is typically plotted in the plane $\mathbb{R}^2$, with each feature as a point above the diagonal $b = d$.
Features with long lifetimes (points far from the diagonal) are often interpreted as meaningful structural signals in the data, while short-lived features (points near the diagonal) are commonly attributed to noise.
Persistence diagrams thus provide a compact and interpretable summary of the multiscale topological properties of the dataset.

\subsection{Persistence Landscapes}

Although persistence diagrams provide a geometric summary of topological features, they are multisets, represented by points on a plane, which makes it challenging to apply classical statistical and machine learning techniques directly.
To address this problem, \citet{bubenik2015statistical} introduced \emph{persistence landscapes}, a functional summary of persistent homology that embeds the information of a persistence diagram into a Banach space, enabling the use of standard statistical tools.

Given a persistence diagram $\mathcal{D} = \{ (b_i, d_i) \}_{i=1}^m$,
we first associate each birth-death pair $(b_i, d_i)$ with a piecewise linear ``tent'' function.

\[
\lambda_{(b,d)}(t) =
\begin{cases}
t-b, & b \leq t \leq \tfrac{b+d}{2}, \\
d-t, & \tfrac{b+d}{2} < t \leq d, \\
0, & \text{otherwise}.
\end{cases}
\]
This function attains its maximum value, $\tfrac{d_i - b_i}{2}$, at the midpoint of the interval.
The persistence landscape is then defined as the sequence of functions

\[
\lambda_k(t)
=
\operatorname{kth\ largest}
\{\lambda_{(b_i,d_i)}(t)\}_{i=1}^m,
k=1,2,\dots
\]

\noindent for each $t \in \mathbb{R}$.
Thus, $\lambda_1$ records the largest ``tent'' value at each $t$, $\lambda_2$ records the second largest, and so forth.
Collectively, the functions $\{\lambda_k\}_{k \geq 1}$ constitute the persistence landscape. This function-space representation is used to define PLDiv in Section~4.

\section{Methodology}

\subsection{Diversity Measure via Persistence Landscapes}

\textbf{Definition 4.1.} Let $\mathcal{X} = \{x_1, \dots, x_n\}$ be a dataset and let $\Lambda(\mathcal{X}) = \{ \lambda_k \}_{k \geq 1}$
denote the persistence landscape constructed from its finite persistence intervals in a fixed homology dimension.
The \emph{persistence landscapes based diversity} score, PLDiv$(\mathcal{X})$, is defined as
\begin{equation}
\mathrm{PLDiv}(\mathcal{X})
= \sum_{k=1}^\infty \int_{\mathbb{R}} \lambda_k(t) \, dt.
\end{equation}
The summation is typically finite, as only a finite number of $\lambda_k$  terms are actually non-zero.
PLDiv$(\mathcal{X})$ measures the cumulative ``area under the triangles'' of the persistence landscape and quantifies the persistence of structural organization in the data. We suppress the homology dimension from the notation; the definition applies separately to each homology dimension, such as $H_0$ and $H_1$ used in this work.

\textbf{Proposition 4.2.} A closed form of PLDiv can be derived. Let $\mathcal{D}=\{(b_i,d_i)\}_{i=1}^m$ be the set of finite birth--death pairs produced by persistence homology, then
\[
\begin{aligned}
\mathrm{PLDiv}(\mathcal{X})= \frac{1}{4}\sum_{i=1}^m (d_i-b_i)^2 .
\end{aligned}
\]

\textit{Sketch of Proof.} $\mathrm{PLDiv}(\mathcal{X}) = \sum_{k=1}^\infty \!\int_{\mathbb{R}} \lambda_k(t)\,dt
 = \sum_{i=1}^m \!\int_{\mathbb{R}} \lambda_{(b_i,d_i)}(t)\,dt.$ Each tent function with its support on the interval $[b_i,d_i]$ is a symmetric isosceles triangle of base length $d_i-b_i$ and height $(d_i-b_i)/2$, hence its area is
\[
\int_{\mathbb{R}} \lambda_{(b_i,d_i)}(t)\,dt \;=\; \tfrac12\cdot (d_i-b_i)\cdot \tfrac{d_i-b_i}{2} \;=\; \frac{(d_i-b_i)^2}{4}.
\]

Summing them yields the closed form above. The quadratic exponent is induced by the area of each landscape tent rather than introduced as a tuned hyperparameter.
A detailed proof is provided in Appendix~\ref{appex:b}.

We next present three remarks that provide intuition for the behavior of PLDiv.

\textbf{Remark 4.3 (Geometric Interpretation).}
The area under $\lambda_k$ summarizes how long the $k$-th most prominent persistence feature remains present as the filtration scale increases. Since birth and death times are determined by distances in the underlying metric space, persistence lifetimes describe how the shape of the dataset evolves as nearby regions progressively merge. Aggregating these lifetimes across $k$ yields a scalar summary of the dataset's structural diversity.

\textbf{Remark 4.4 (Interpreting PLDiv Values).}
A large PLDiv$(\mathcal{X})$ indicates that persistent structural features remain present over a wide range of filtration values, suggesting substantial structural diversity in the dataset. In contrast, a small PLDiv$(\mathcal{X})$ indicates that most features disappear quickly as the filtration grows, reflecting a more homogeneous or structurally concentrated representation.

\textbf{Remark 4.5 (Statistical View).}
The persistence landscape represents the finite birth--death pairs as a sequence of functions, and PLDiv$(\mathcal{X})$ summarizes this representation by the total area under all landscape layers.
By Proposition~4.2, PLDiv$(\mathcal{X})$ equals one quarter of the sum of squared persistence lifetimes. This area-based summary jointly reflects the number and magnitude of nonzero persistent features while assigning greater weight to longer-lived structure.

\subsection{Axiomatic Properties of Diversity}

Among core diversity axiomatic properties provided by \citet{leinster2012measuring} and  \citet{leinster2021entropy}, our proposed diversity measure, PLDiv, satisfies four fundamental axioms: effective size, monotonicity, twin property, and symmetry. We describe each property below and provide formal proofs in Appendix~\ref{app:axiom}; comparisons with existing methods appear in Appendix~\ref{app:apc}. These axioms provide consistency criteria rather than guarantees of practical effectiveness, while their violation may indicate undesirable behavior.

\begin{itemize}

    \item \textbf{Effective size.}
    For a fixed number of points, PLDiv$(\mathcal{X})$ reflects the effective spread of the dataset: it is larger when the data points are more separated and smaller when they are more concentrated. In particular, the $0$-dimensional PLDiv satisfies
    \[
    0\leq\mathrm{PLDiv}(\mathcal{X})\leq\frac{(n-1)c^2}{4}.
    \]
    where $c$ is the maximum pairwise distance between points in $\mathcal{X}$. The lower bound is attained when all points coincide. Illustrations are provided in Section \ref{exp:text} and Appendix~\ref{appendix:tail}.

    \item \textbf{Monotonicity.}
    Decreasing similarity increases diversity. For a fixed sample size, if pairwise similarity is decreased through a uniform dilation of the underlying metric $D$ to $\alpha D$, then every finite persistence interval \((b_i,d_i)\) becomes \((\alpha b_i,\alpha d_i)\) and
    \[
    \mathrm{PLDiv}(\alpha \mathcal{X})=\alpha^{2}\,\mathrm{PLDiv}(\mathcal{X}).
    \]

    Hence, when $\alpha>1$ and $\mathrm{PLDiv}(\mathcal{X})>0$, the score strictly increases; when $0<\alpha<1$, it strictly decreases.

    \item \textbf{Twin property.} Adding an exact duplicate of a point does not change PLDiv$(\mathcal{X})$.
    The duplicate induces a trivial birth--death pair $(0,0)$, contributing zero to the diversity score.
    Let $\mathcal{X}$ be a dataset and let $x_i \in \mathcal{X}$. For the set $\mathcal{X}' = \mathcal{X} \cup \{x_n\}$ where $x_n = x_i$, the diversity is unchanged:
    \[
    \mathrm{PLDiv}(\mathcal{X}') = \mathrm{PLDiv}(\mathcal{X}).
    \]

    \item \textbf{Symmetry.} PLDiv is invariant to the ordering of data points (permutation invariance).
    Since persistent homology depends only on the metric structure of $\mathcal{X}$ and PLDiv$(\mathcal{X})$ is computed from the multiset of intervals $\{(b_i,d_i)\}$, relabeling or reordering points does not affect the value of the score.
    Let $\mathcal{X}=(x_1,\dots,x_n)$ be an ordered sequence of points and let $\pi$ be any permutation of $\{1,\dots,n\}$. For the permuted sequence $\mathcal{X}_\pi=(x_{\pi(1)},\dots,x_{\pi(n)})$, we have
    \[
    \mathrm{PLDiv}(\mathcal{X}_\pi) = \mathrm{PLDiv}(\mathcal{X}).
    \]

\end{itemize}

\subsection{Scalable Sparse Implementation}

The definition and closed form above apply to persistence diagrams obtained using any filtration and homology dimension. In our experiments, PLDiv is instantiated using the Vietoris-Rips filtration. We use 0-dimensional persistent homology ($H_0$), which tracks connected components, throughout the empirical evaluation; we also use 1-dimensional persistent homology ($H_1$) in the reasoning-trace homogenization experiments to assess the additional value of loop-sensitive topology.

The $H_0$ instantiation enables a scalable sparse implementation. Under the $H_0$ Vietoris--Rips filtration, the finite death times coincide with the edge weights of the minimum spanning tree (MST) of the complete metric graph \citep{hofer2020topologically}, and therefore
\[
\mathrm{PLDiv}(\mathcal{X})=\frac14\sum_{e\in\mathrm{MST}}w(e)^2.
\]
We use this identity as a computational shortcut: a symmetrized $k$-nearest-neighbor graph is constructed with scikit-learn and its sparse MST is computed with SciPy \citep{pedregosa2011scikit,virtanen2020scipy}, avoiding materialization of the dense complete graph. The formal estimator, its exactness condition, and empirical validation are provided in Appendix~\ref{app:knn-sparse-pldiv} and Tables~\ref{tab:knn-sparse-k-ablation} and~\ref{tab:fair-runtime-three-metrics}.

\section{Experiment}

\subsection{Semantic Diversity in Text Embeddings}
\label{exp:text}

\begin{figure*}[!ht]
    \centering
    \includegraphics[width=1.0\linewidth]{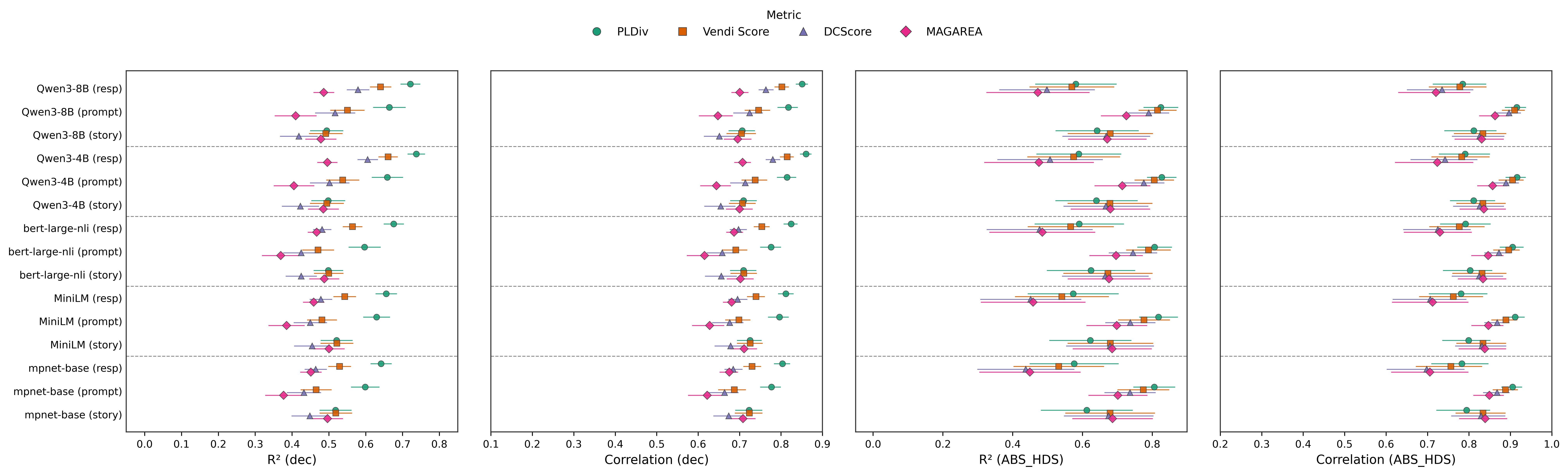}
    \caption{\textbf{Comparison of four diversity metrics for predicting the temperature proxy and human diversity judgments across tasks and embedding models.} This figure uses Euclidean distances for PLDiv and MagArea and RBF kernels for Vendi Score and DCScore. Points with different shapes denote different metrics. For $R^2$, error bars indicate the standard deviation of repeated five-fold cross-validation scores; for Pearson correlation, they indicate row-level bootstrap 95\% confidence intervals from 1,000 resamples. Experiments with \textit{ABS-HDS} exhibit larger error bars due to its smaller sample size. Detailed MSE results are provided in Fig.~\ref{fig:text_mse}, and additional results using different distance matrices and similarity kernels are shown in Fig.~\ref{fig:cs_eu}, both in Appendix~\ref{appedix:text}.}
    \label{fig:text_emb}
\end{figure*}

We evaluate PLDiv for measuring semantic diversity using the dataset from \citet{tevet-berant-2021-evaluating}, which includes three generation tasks: story completion (story), dialogue response generation (resp), and three-word completion (prompt). Each task contains 1,000 input contexts, each paired with a response set of 10 candidate outputs generated at a fixed softmax temperature $dec$. Different response sets are generated at different temperature values. In addition, each task includes a human-evaluated subset of 200 input contexts, each with 10 candidate outputs. Each response set is assigned a mean human diversity score (\textit{ABS-HDS}) based on human annotations. We use the decoding temperature $dec$ as a proxy for generation diversity and \textit{ABS-HDS}, a five-point Likert-scale human diversity score averaged across annotators, as the human reference.

Accordingly, we use linear regression with 5-fold cross-validation to evaluate how well diversity metrics capture variation in (dec) and \textit{ABS-HDS}. Performance is assessed using $R^2$ and MSE. In addition, we compute Pearson correlations and perform 1,000 bootstrap iterations to estimate confidence intervals. Each response set is embedded using five models:
\texttt{\seqsplit{bert-large-nli-stsb-mean-tokens}},
\texttt{\seqsplit{all-MiniLM-L12-v2}},
\texttt{\seqsplit{all-mpnet-base-v2}},
\texttt{\seqsplit{Qwen3-Embedding-4B}},
and
\texttt{\seqsplit{Qwen3-Embedding-8B}}.

Figure \ref{fig:text_emb} summarizes the $R^2$ and Pearson correlation results across tasks and embedding models. PLDiv performs consistently well for both the temperature-based diversity proxy and human \textit{ABS-HDS} judgments on prompt completion and dialogue response generation. The main exception is story completion under \textit{ABS-HDS}, where task-specific properties are not captured equally by different diversity metrics. This task-specific difference motivates a closer examination of exact response repetition.

To better understand the different behavior on story completion, we examined exact response duplication within each ten-response set. A post-hoc audit shows that exact duplication is substantially more prevalent in story completion, affecting 58.1\% of its response sets, and that story \textit{ABS-HDS} is strongly associated with the number of unique responses.

This distinction is relevant because PLDiv satisfies the \emph{twin property}: an exact copy contributes only a zero $H_0$ lifetime. PLDiv therefore responds to the loss of distinct semantic support, but is insensitive to repeated occurrences of an already represented response. In contrast, Vendi Score and DCScore more directly reflect exact duplication. Consistent with this distinction, after removing exact duplicates, PLDiv achieves higher correlations with human judgments across embedding models. Detailed results of this deduplication analysis are reported in Appendix~\ref{app:text-duplicate-sensitivity}.

\subsection{Data Near-Duplicate Collapse Detection}
\label{sec:instruction-collapse}

\begin{figure}[t]
    \centering
    \includegraphics[width=0.6\linewidth]{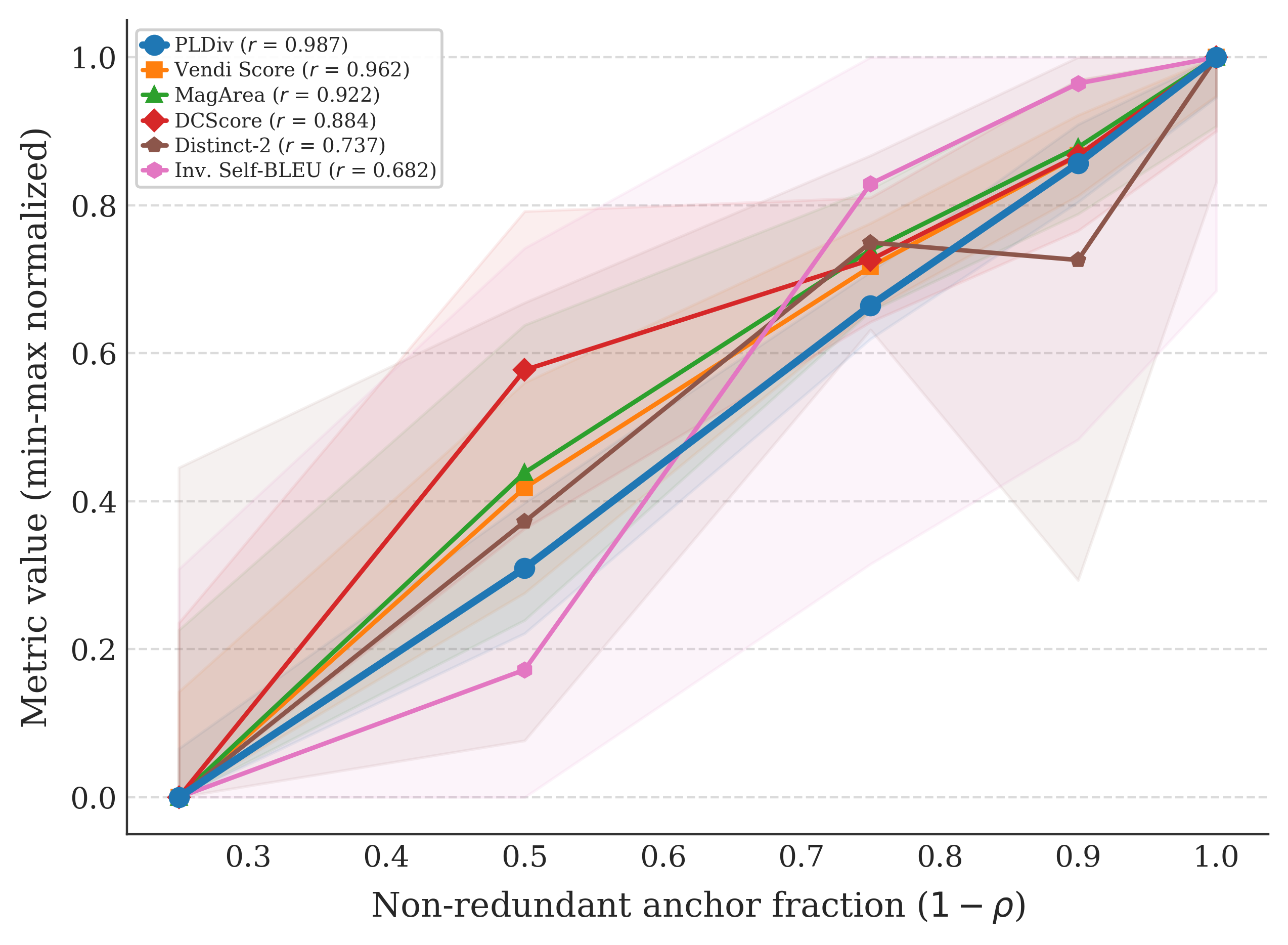}
    \caption{\textbf{Instruction-data near-duplicate collapse.} As the replacement fraction $\rho$ increases, more examples are replaced with nearest-neighbor variants of retained anchors. PLDiv shows the highest Pearson correlation with the remaining non-redundant fraction among the existing metrics, as reported in the legend.}
    \label{fig:instruction-collapse}
\end{figure}

To further evaluate PLDiv in dataset construction and curation settings, we use Databricks Dolly-15K \citep{conover2023dolly} to construct a controlled near-duplicate collapse instruction-following benchmark, where the dataset size and task-category distribution remain fixed while semantic redundancy is progressively increased. Each instruction is embedded using \texttt{\seqsplit{all-mpnet-base-v2}}. To account for sampling variability, we repeat the construction process over 10 independent trials and report mean values and standard deviations.

For each trial, we first sample a balanced anchor set $S_0$ containing $N=400$ instructions across task categories. We then define a collapse level $\rho \in \{0,0.1,0.25,0.5,0.75\}$, which controls the proportion of examples replaced by near-duplicate variants. Specifically, a fraction $\rho$ of the dataset is replaced with nearest-neighbor instructions of the retained anchors, selected from the same task category but outside $S_0$ using cosine similarity on embeddings. The remaining $(1-\rho)N$ instructions are kept unchanged. This procedure preserves dataset size and category composition while progressively reducing semantic independence within the dataset, simulating realistic near-duplicate collapse in instruction-tuning datasets. Table~\ref{tab:instruction-collapse-diagnostics} in Appendix \ref{app:instruction-collapse} reports the mean cosine similarity between retained anchors and their nearest-neighbor replacements, compared with randomly selected same-category examples. The substantially higher similarity of anchor--neighbor pairs confirms that the constructed datasets exhibit controlled local redundancy.

We compare PLDiv against Vendi Score, DCScore, and MagArea, as well as lexical NLG baselines including Distinct-1/2 and inverted Self-BLEU. The evaluation target is the non-redundant fraction $1-\rho$, with stronger diversity metrics expected to increase monotonically as $1-\rho$ increases. Figure~\ref{fig:instruction-collapse} shows that PLDiv achieves the strongest Pearson correlation with the non-redundant fraction among the evaluated metrics, suggesting that it closely tracks changes in dataset diversity and may be useful for monitoring diversity collapse during dataset curation.

\subsection{Homogenization in Reasoning Traces}
\label{sec:h1-reasoning-homogenization}

Chain-of-thought (CoT) supervision has become a central paradigm for training large language models, as it provides not only final answers but also intermediate reasoning traces that shape how models generalize and solve complex problems \cite{wei2022chain, kojimat2022, lightman2023letsverify}.
Many problems admit multiple valid reasoning paths that lead to the same correct solution, implying that reasoning is a structured space of alternative trajectories that traverse distinct intermediate reasoning states before converging on the same answer.
Consequently, preserving this diversity of reasoning trajectories is fundamental to learning robust reasoning behaviors. Ignoring such diversity collapses rich supervisory signals into a single canonical reasoning pattern, giving rise to \emph{reasoning collapse}, a failure mode in which structurally distinct solution paths are mapped to overly similar intermediate representations, ultimately limiting both interpretability and the model's ability to generalize.

We employ ProcessBench \citep{zheng2024processbench} to construct controlled reasoning trajectories for evaluating PLDiv, selecting problems that contain at least two independently generated reasoning trajectories leading to the same correct final answer. After quality filtering, the resulting benchmark contains 90 reasoning pairs (88 from OlympiadBench and 2 from GSM8K), where one trajectory is designated as the prototype and the other as the alternative reasoning path. Rather than treating each chain of thought as a single textual sequence, we represent it as an ordered sequence of intermediate reasoning states. This state-level representation preserves the temporal progression of reasoning and enables fine-grained manipulation of individual reasoning steps while keeping the surrounding reasoning context unchanged.

\begin{figure}[H]
    \centering
    \includegraphics[width=0.98\linewidth]{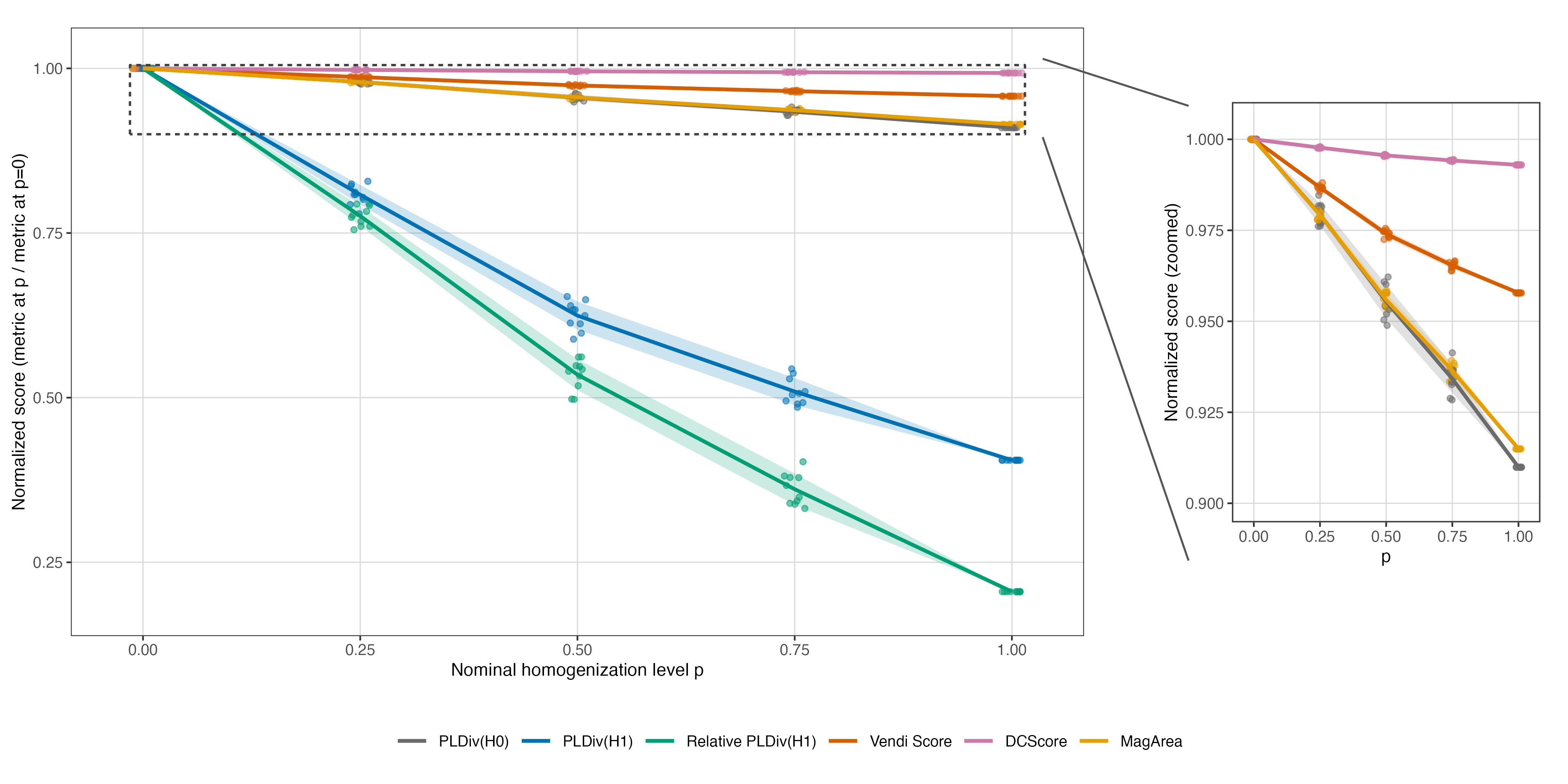}
    \caption{\textbf{Sensitivity of diversity metrics to reasoning-strategy collapse.} Every structural metric receives the same normalized trajectory-graph distance matrix. The dashed rectangle marks the low-variation range from $0.90$ to $1.005$, and the zoomed-in panel shows $\mathrm{PLDiv}(H_0)$, Vendi Score, DCScore, and MagArea within this range. Within each replacement-order trial, each metric is divided by its value at $p=0$. Small dots show the 10 individual trial values, solid lines show their means, and shaded intervals show the mean $\pm1$ sample standard deviation. Points overlap at $p=0$ and $p=1$ because these endpoint conditions are invariant to replacement order. A steeper decline indicates greater sensitivity to the collapse intervention, while a flatter curve indicates weaker sensitivity.}
    \label{fig:reasoning_graph_curves}
\end{figure}

To simulate reasoning collapse, we define five homogenization levels, $p \in \{0,0.25,0.5,0.75,\allowbreak 1.0\}$. At each level, we randomly replace a fraction $p$ of the alternative trajectory with semantically equivalent paraphrases of the aligned prototype steps. All paraphrases are generated and validated by GPT-5.6 for semantic and mathematical equivalence. Throughout this process, the reasoning trajectories, temporal ordering, and shared question--answer endpoints remain unchanged; only the textual realization of selected intermediate reasoning states is modified. Consequently, the benchmark isolates the effect of intermediate-state homogenization from changes in the overall reasoning process. Full construction details are provided in Appendix~\ref{app:h1-reasoning-homogenization}.

To evaluate the relative contribution of cyclic reasoning paths from the global geometric scale of the representation, we define the \textbf{Relative $\text{PLDiv}(H_1)$}:
\[
\mathrm{PLDiv}_{rel}(H_1) = \frac{\mathrm{PLDiv}(H_1)}{\mathrm{PLDiv}(H_0) + \mathrm{PLDiv}(H_1)}
\]
where $\mathrm{PLDiv}(H_0)$ and $\mathrm{PLDiv}(H_1)$ denote the 0- and 1-dimensional PLDiv, respectively. While $H_0$ quantifies the global trajectory dispersion, $H_1$ isolates the cyclic topology formed by divergent reasoning paths.

Figure~\ref{fig:reasoning_graph_curves} visualizes the metric response trajectories under increasing reasoning collapse, whose endpoint changes correspond directly to the $\Delta$ values reported in Table~\ref{tab:h1-summary}; the table further evaluates each metric's ability to recover the realized homogenization level through problem-grouped out-of-fold prediction across 10 independent trials. In multi-step reasoning, distinct solution paths typically share the same problem statement and final answer while diverging at intermediate steps, naturally forming closed metric loops. In this regime, existing paradigms inherently struggle: lexical metrics are deceived by superficial paraphrase distinctions, while distance- and entropy-based metrics ($\mathrm{PLDiv}(H_0)$, Vendi Score, DCScore, MagArea)  primarily capture spatial dispersion. In contrast, $\mathrm{PLDiv}(H_1)$ directly captures the geometric closure of these alternative reasoning loops. Both $\mathrm{PLDiv}(H_1)$ and $\mathrm{PLDiv}_{rel}(H_1)$ achieve superior sensitivity and out-of-fold performance in tracking reasoning collapse, demonstrating that
higher-dimensional persistence adds informative geometric signal beyond global dispersion when reasoning
diversity is expressed through branching and multi-path structure.

\begin{table}[t]
\centering
\small
\setlength{\abovecaptionskip}{\baselineskip}
\setlength{\belowcaptionskip}{\baselineskip}
\caption{\textbf{Comparison of reasoning homogenization and predictive power across diversity metrics.} Every structural metric receives the same trajectory-graph distance matrix. $\Delta$ (\%) is the relative percentage change from $p=0$ to $p=1$. \textit{Concord.} counts problems whose trial-averaged change has the same direction as the overall change across all problems. High concordance indicates that the overall trend is consistent across individual problems. $R^2$ and RMSE are obtained from 10-fold cross-validation and reported as mean $\pm$ standard deviation across 10 trials.}

\label{tab:h1-summary}
\begin{tabular}{lcccc}
\toprule
\textbf{Metric} &
\textbf{$\Delta (\%)$} &
\textbf{Concord. ($\uparrow$)} &
\textbf{$R^2$ ($\uparrow$)} &
\textbf{RMSE ($\downarrow$)} \\
\midrule

Vendi Score & -4.22 & 88/90 &
$0.320 \pm 0.009$ &
$0.293 \pm 0.002$ \\

DCScore & -0.70 & 87/90 &
$0.252 \pm 0.013$ &
$0.307 \pm 0.003$ \\

MagArea & -8.51 & 86/90 &
$0.218 \pm 0.004$ &
$0.314 \pm 0.001$ \\

Distinct-1 & 0.47 & 52/90 &
$0.00 \pm 0.00$ &
$0.36 \pm 0.00$ \\

Distinct-2 & -6.27 & 53/90 &
$0.05 \pm 0.00$ &
$0.35 \pm 0.00$ \\

Inv.Self-BLEU & -17.87 & 66/90 &
$0.11 \pm 0.00$ &
$0.34 \pm 0.00$ \\

\midrule

$\mathrm{PLDiv}(H_0)$ & -9.01 & 85/90 &
$0.212 \pm 0.008$ &
$0.315 \pm 0.002$ \\

$\mathbf{\mathrm{PLDiv}}(H_1)$ & $\mathbf{-59.50}$ & $\mathbf{86/90}$ &
$\mathbf{0.556 \pm 0.014}$ &
$\mathbf{0.237 \pm 0.004}$ \\

$\mathbf{\mathrm{PLDiv}}_{\mathrm{rel}}(H_1)$ & $\mathbf{-79.46}$ & $\mathbf{89/90}$ &
$\mathbf{0.557 \pm 0.014}$ &
$\mathbf{0.236 \pm 0.004}$ \\

\bottomrule
\end{tabular}
\end{table}

\subsection{Diversity in Image Embeddings}
To evaluate the applicability of PLDiv beyond NLP settings, we test it on the Colored MNIST image dataset \citep{deng2012mnist}. Following \citet{ospanov2024towards}, the number of labels is used as a proxy for dataset diversity, with larger label classes corresponding to more diverse image sets. In this experimental setup, the total dataset size is fixed at 750 samples. As the number of classes increases, samples from existing classes are proportionally replaced by samples from newly introduced classes, resulting in progressively more diverse class compositions while maintaining a constant dataset size. For each class count, both the initial class and the newly introduced classes are randomly selected, and the experiment is repeated over 10 independent trials. This setting reflects practical dataset curation scenarios where the overall annotation budget remains fixed while semantic coverage is expanded over time. Accordingly, we evaluate whether PLDiv can reliably capture changes in dataset diversity that arise solely from differences in class composition rather than dataset size.

\begin{figure}[!h]
    \centering
        \includegraphics[width=0.49\linewidth]{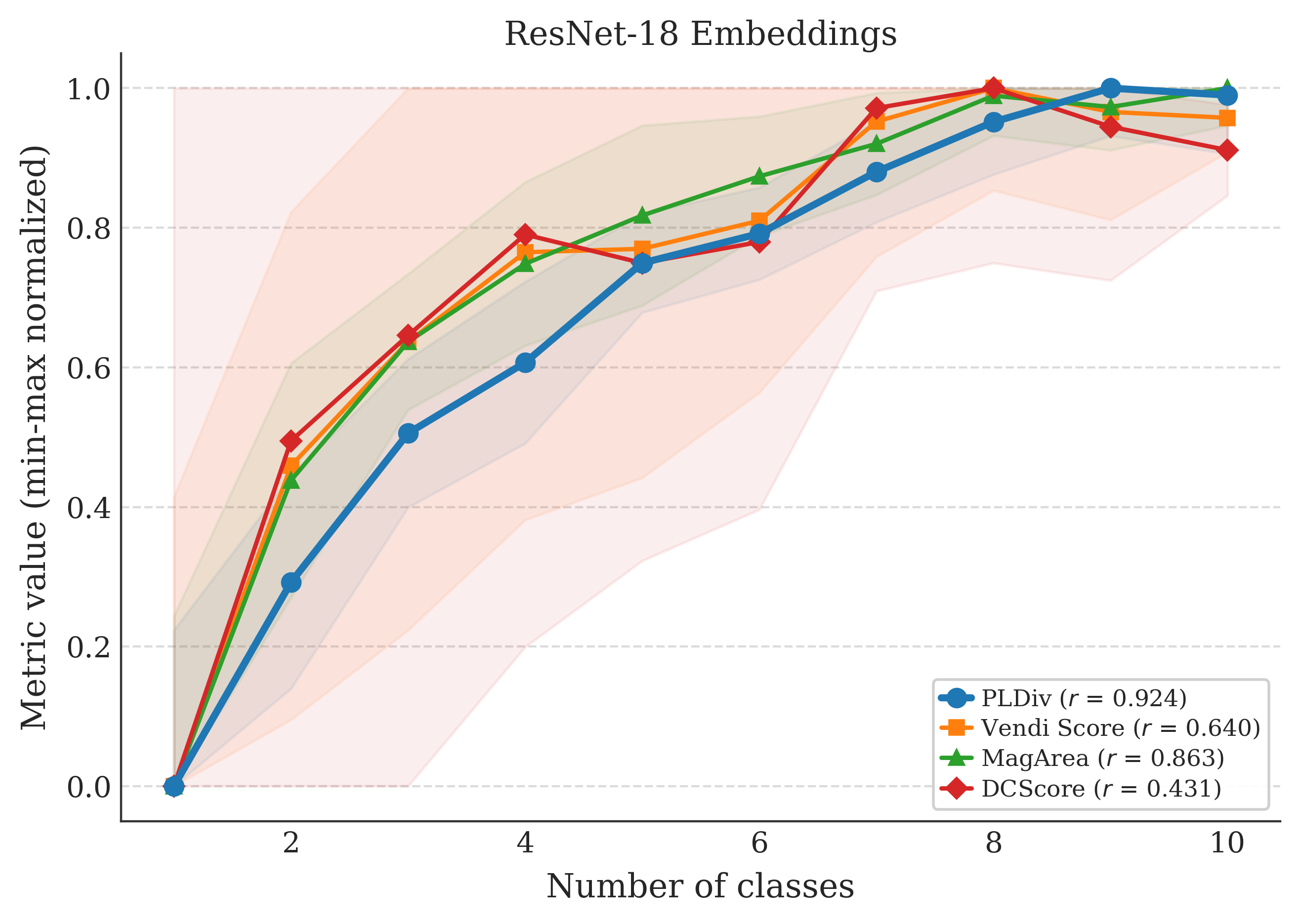}
        \includegraphics[width=0.49\linewidth]{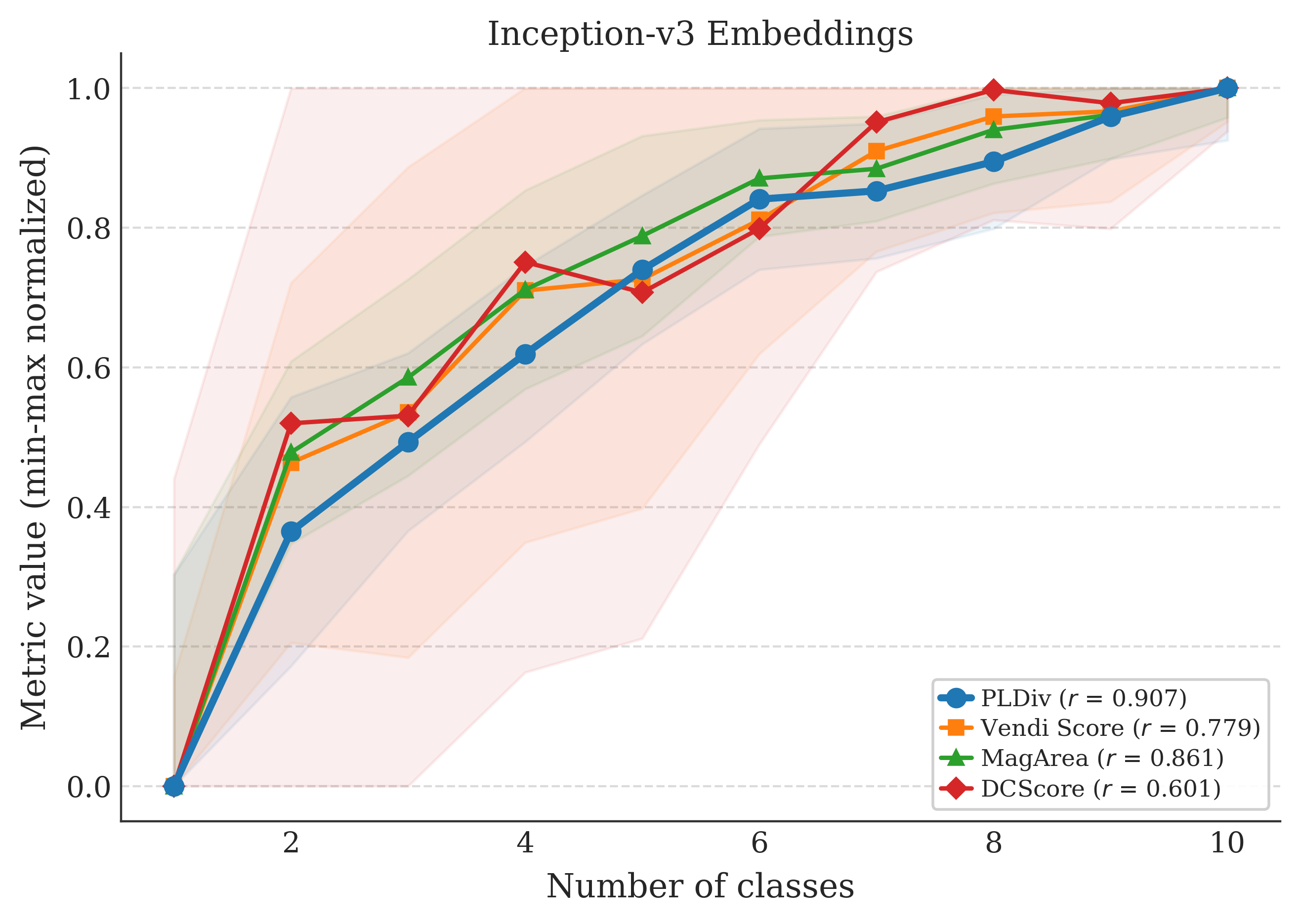}
 \caption{\textbf{PLDiv correlates strongly with class diversity and shows consistent trends across two image embedding models, while competing baselines exhibit lower or less consistent correlations.}}
\label{fig:image}
\end{figure}

Metrics are computed on image embeddings rather than raw pixels. We use Inception V3 and ResNet-18 to extract 2048- and 512-dimensional embeddings, respectively. Figure~\ref{fig:image} shows that PLDiv achieves the highest Pearson correlation with class diversity and consistently outperforms the evaluated baseline metrics across both embedding models. This provides evidence that PLDiv is not tied to a single image embedding geometry. These results support the applicability of PLDiv to vision datasets represented in semantic embedding spaces.

\section{Conclusion}

In this work, we introduced PLDiv, a persistence landscapes-based metric for measuring dataset diversity and semantic collapse from the underlying structural organization of data. By integrating the total area of persistence landscapes, PLDiv provides an interpretable scalar summary that reflects both the number and persistence of structural features. Its closed form, axiomatic properties, and applicability across homology dimensions provide a unified basis for analyzing component-level coverage and higher-order structural organization.

Empirically, PLDiv captures two complementary patterns: the expansion of meaningful coverage and the contraction of data toward redundant or homogenized configurations. It tracks diversity changes in text generation and image-class coverage, while detecting collapse caused by near-duplicate instructions and homogenized reasoning trajectories. The controlled PCA experiment further demonstrates a strong association with intrinsic dimensionality in both raw and trace-normalized settings (Figures~\ref{fig:pca-raw} and~\ref{fig:pca-trace-norm} in Appendix~\ref{tab:pca-dim}). Together, these results highlight the potential of PLDiv as an exploratory data analysis tool for examining structural complexity, semantic coverage, and collapse. Its scalable sparse implementation further suggests its applicability in large-scale data-centric workflows, including synthetic-data monitoring, dataset curation, and generative-model evaluation.

\subsubsection*{Broader Impact Statement}

PLDiv is intended as a diagnostic tool for monitoring structural diversity
and semantic collapse in datasets. It supports more transparent dataset
curation by identifying redundancy and loss of semantic coverage. However,
PLDiv measures diversity relative to a chosen representation and distance
function; it does not measure data quality, factuality, fairness, demographic
representation, or the social value of individual examples. Biases in the
embedding model or a mismatch between the representation and the target domain
may therefore affect its scores. We recommend using PLDiv only as one
diagnostic signal alongside task-specific quality, safety, privacy, and
subgroup analyses, with human review for consequential dataset-curation
decisions.

Our experiments use previously released datasets and do not recruit new
participants or collect new personal data. We reuse previously released
ABS-HDS annotations, and GPT-5.6 is used only to generate and validate
controlled paraphrases for the reasoning-collapse experiment. Because
automated paraphrase validation may not detect all errors or biases, the
generated material is used only for controlled evaluation rather than
deployment.

\bibliography{refs}
\bibliographystyle{tmlr}

\appendix

\section{Description of Diversity Scores in Comparisons}
\label{app:aa}

Vendi Score (VS) \citep{dan2023vendi}, derived from a set of samples and their pairwise similarity functions, quantifies the similarities among the data in a dataset.
Mathematically, VS is given by the exponential of the Shannon entropy, which is obtained from the eigenvalues of the scaled similarity matrix \( X^\top X \):
\[
VS
= \exp \left( - \sum_{i=1}^{n} \lambda_i \log \lambda_i \right)
\]
where \( \lambda_i \) are the eigenvalues of scaled \( X^\top X \).

\citet{limbeck2024metric} introduces several \emph{magnitude-based} diversity
measures that leverage the notion of the effective size of a metric space across scales. The core idea
is to compute the \emph{magnitude function}, $\text{Mag}_X(t)$, which tracks how the effective
number of points in a space changes as pairwise distances are rescaled. To summarise this behaviour,
the authors propose two derived metrics: the area under the magnitude function (MagArea) as a
reference-free measure of intrinsic diversity, and the difference between magnitude functions
(MagDiff) as a reference-based measure:
\[
\begin{aligned}
\text{MagArea}
&= \int_{t_0}^{t_{\text{cut}}} \text{Mag}_X(t)\, dt,\\
\text{MagDiff}
&= \int_{t_0}^{t_{\text{cut}}}
\big(\text{Mag}_X(t)-\text{Mag}_Y(t)\big)\, dt,
\end{aligned}
\]
where $\text{Mag}_X(t)$ is the magnitude function of $X$ at scale $t$ and $t_{\text{cut}}$ denotes
the convergence scale used for evaluation. These measures provide robust multi-scale summaries of diversity and have been shown to detect phenomena such as curvature, mode collapse, and mode dropping in text, image, and graph representations.

\citet{zhu2025measuring} proposes \textbf{DCScore}, which departs from entropy or scale-based approaches by reframing diversity measurement as a \textit{classification problem}. Instead of relying on eigenvalue decomposition or scale-sensitive geometric measures, DCScore evaluates how well each individual sample in a dataset can be distinguished from all others. Specifically, each sample is treated as its own class, and pairwise similarities are converted into classification probabilities through a softmax function. The resulting score is then defined as the trace of the resulting probability matrix:
\[
\begin{aligned}
\text{DCScore}(D)
&= \mathrm{tr}(P)=\sum_{i=1}^n P[i,i],\\
P[i,j]
&= \frac{\exp\left(K[i,j]/\tau\right)}
{\sum_{k=1}^n \exp\left(K[i,k]/\tau\right)},
\end{aligned}
\]
where \(K[i,j]\) denotes the similarity between samples \(i\) and \(j\), and \(\tau\) is a temperature parameter that controls the classification sharpness. This formulation is principled and efficient, emphasizing sample separability without considering the geometric or topological structure of the dataset, which can also be important for characterizing diversity.

For lexical NLG baselines, we include type-token dispersion and leave-one-out overlap metrics. Given a collection of generated texts \( C = \{ y_i \}_{i=1}^N \), Distinct-\( n \) is
\[
\mathrm{Distinct}\text{-}n(C)
=
\frac{\left|\bigcup_{i=1}^{N} \mathrm{ngrams}_n(y_i) \right|}
{\sum_{i=1}^{N} \left| \mathrm{ngrams}_n(y_i) \right|},
\]
where \( n = 1 \) or \( 2 \) in our experiments \citep{li-etal-2016-diversity,song2024scaling}. Self-BLEU treats each output as a hypothesis and the remaining outputs as references \citep{zhu2018texygen,shu2019generating}:
\[
\mathrm{Self\text{-}BLEU}_4(C)
=
\frac{1}{N}
\sum_{i=1}^N
\mathrm{BLEU}_4 \big(y_i, C \setminus \{ y_i \} \big).
\]
Because lower Self-BLEU indicates higher diversity, we report inverse Self-BLEU as
\[
\mathrm{Inv.\,Self\text{-}BLEU}(C) = 1 - \mathrm{Self\text{-}BLEU}_4(C).
\]

\section{Mathematical Proofs}
\label{appex:b}

\subsection{PLDiv Closed Form }
Let $\mathcal D=\{(b_i,d_i)\}_{i=1}^m$ be a finite multiset of persistence birth--death pairs and let $\lambda_{(b_i,d_i)}:\mathbb R\to[0,\infty)$ denote the usual persistence ``tent'' function associated to the interval $(b_i,d_i)$. Let $\{\lambda_k(t)\}_{k\ge1}$ be the persistence landscape functions obtained by ordering the values $\{\lambda_{(b_i,d_i)}(t)\}_{i=1}^m$ at each fixed $t$ in nonincreasing order (with $\lambda_k(t)=0$ for all $k>m$). Then

\[
\begin{aligned}
\mathrm{PLDiv}(\mathcal{X})
&= \sum_{k=1}^\infty \int_{\mathbb{R}} \lambda_k(t)\,dt\\
&= \sum_{i=1}^m \int_{\mathbb{R}} \lambda_{(b_i,d_i)}(t)\,dt\\
&= \frac{1}{4}\sum_{i=1}^m (d_i-b_i)^2.
\end{aligned}
\]

\begin{proof}
By definition $\lambda_k(t)$ are the order statistics (at each fixed $t$) of the family $\{\lambda_{(b_i,d_i)}(t)\}_{i=1}^m$. For any finite collection of nonnegative functions $f_i(t)$,
\[
\sum_{k=1}^\infty \text{$k$-th largest of $\{f_i(t)\}$} \;=\; \sum_{i=1}^m f_i(t),
\]

Applying this pointwise gives
\[
\sum_{k=1}^\infty \lambda_k(t) = \sum_{i=1}^m \lambda_{(b_i,d_i)}(t).
\]

Each $\lambda_{(b_i,d_i)}$ is continuous with compact support $[b_i,d_i]$, hence measurable and integrable. Since $\mathbb{R}$ with Lebesgue measure and $\mathbb{N}$ with counting measure are $\sigma$-finite, and all landscape functions are nonnegative and measurable, the hypotheses of Tonelli's theorem are satisfied. By Tonelli's theorem \citep{tao2011introduction},
\[
\begin{aligned}
\sum_{k=1}^\infty \int_{\mathbb R} \lambda_k(t)\, dt
&= \int_{\mathbb R} \sum_{k=1}^\infty \lambda_k(t)\, dt\\
&= \int_{\mathbb R} \sum_{i=1}^m \lambda_{(b_i,d_i)}(t)\, dt\\
&= \sum_{i=1}^m \int_{\mathbb R} \lambda_{(b_i,d_i)}(t)\, dt.
\end{aligned}
\]

Finally, each tent function supported on the interval $[b_i,d_i]$ is a symmetric isosceles triangle of base length $d_i-b_i$ and height $(d_i-b_i)/2$, hence its area is
\[
\begin{aligned}
\int_{\mathbb{R}} \lambda_{(b_i,d_i)}(t)\,dt
&= \tfrac12\cdot (d_i-b_i)\cdot \tfrac{d_i-b_i}{2}\\
&= \frac{(d_i-b_i)^2}{4},
\end{aligned}
\]
Summing over $i=1,\dots,m$ gives the final identity
\[
\sum_{i=1}^{m} \int_{\mathbb{R}} \lambda_{(b_i,d_i)}(t)\, dt
= \frac{1}{4} \sum_{i=1}^{m} (d_i - b_i)^2.
\]

\end{proof}

\subsection{Axiomatic Properties of Diversity}
\label{app:axiom}

A diversity measure derived from Persistence Landscapes (PLs) is defined as a summary statistic of the persistence lifetimes generated from a dataset's Vietoris-Rips filtration. We prove that such a measure satisfies the key principles of effective size, monotonicity, the twin property, and symmetry.

\begin{itemize}

    \item \textbf{Effective size.}
    For a fixed number of points, PLDiv$(\mathcal{X})$ reflects the effective spread of the dataset: it is larger when the data points are more separated and smaller when they are more concentrated.

 \begin{proof}

$\mathcal{X} = \{x_1, \dots, x_n\}$ is a point cloud of $n$ points in a metric space $(M, D)$. Let
\[
c = \max_{i \ne j} D(x_i, x_j)
\]
be the maximum pairwise distance of the point cloud. In the Vietoris--Rips filtration, all $n$ points are born at radius $\epsilon = 0$ (i.e., $b_i = 0$ for all $i$). There will be $n-1$ merging events with death times
\[
d_1, d_2, \dots, d_{n-1}.
\]

Suppose there exists a configuration of $n$ points with maximum pairwise distance $c$ such that its PLDiv exceeds $\frac{n-1}{4}c^2$:
\[
\begin{aligned}
\operatorname{PLDiv}(\mathcal{X})
&=
\frac{1}{4}\sum_{i=1}^{n-1}(d_i-0)^2\\
&=
\frac{1}{4}\sum_{i=1}^{n-1} d_i^2\\
&>
\frac{n-1}{4}c^2.
\end{aligned}
\]
Multiplying both sides by $4$ gives
\[
\sum_{i=1}^{n-1} d_i^2 > (n-1)c^2.
\]

Since the sum of the $n-1$ squared terms is strictly greater than $(n-1)c^2$, the average of these squared terms must also be greater than $c^2$. Consequently, there must exist at least one death time $d_k$ such that
\[
d_k^2 > c^2,
\]
which implies
\[
d_k > c.
\]

However, this contradicts the definition
\[
c = \max_{i \ne j} D(x_i, x_j).
\]
Therefore, the initial assumption must be false, and the upper bound holds:
\[
\operatorname{PLDiv}(\mathcal{X})
\le
\frac{n-1}{4}c^2.
\]

Now we prove that when all points are distinct and equidistant,
\[
D(x_i,x_j)=c
\qquad \forall i \neq j,
\]
the quantity $\operatorname{PLDiv}$ attains the upper bound.

\paragraph{Sufficient condition.}

Assume that:

\begin{enumerate}
    \item All points are born at $\epsilon = 0$;
    \item For $0 < \epsilon < c$, no merging occurs;
    \item At $\epsilon = c$, all edges are added simultaneously, and all $n-1$ connected components merge at once.
\end{enumerate}

Therefore, all death times satisfy
\[
d_i = c,
\qquad i=1,\dots,n-1.
\]
Substituting into the definition of $\operatorname{PLDiv}$ gives
\[
\operatorname{PLDiv}(\mathcal{X})
=
\frac{1}{4}\sum_{i=1}^{n-1} c^2
=
\frac{n-1}{4}c^2.
\]

\paragraph{Necessary condition.}

\[
\begin{aligned}
&\operatorname{PLDiv}(\mathcal{X})
= \frac{n-1}{4}c^2,\\
&\Longrightarrow\quad
\frac14\sum_{i=1}^{n-1} d_i^2
= \frac{n-1}{4}c^2,\\
&\Longrightarrow\quad
\sum_{i=1}^{n-1} d_i^2
= (n-1)c^2.
\end{aligned}
\]

Hence,
\[
\sum_{i=1}^{n-1}(c^2-d_i^2)=0.
\]

Since
\[
c=\max_{i\ne j}D(x_i,x_j),
\]
we have
\[
c^2-d_i^2 \ge 0
\qquad \forall i.
\]

Because the sum of nonnegative terms equals zero, each term must individually vanish:
\[
c^2-d_i^2=0
\qquad \forall i,
\]
hence
\[
d_i=c
\qquad \forall i\in\{1,\dots,n-1\}.
\]

That is, every merging event occurs exactly at $\epsilon = c$. In the Vietoris--Rips filtration, connected components merge in increasing order of pairwise distances. If there existed any pair of points with distance strictly less than $c$, then the first merging event would occur at that smaller distance, contradicting the fact that all $d_i=c$.

Therefore, no pairwise distance can be strictly smaller than $c$. Since by definition
\[
D(x_i,x_j)\le c,
\]
we conclude that
\[
D(x_i,x_j)=c
\qquad \forall i\neq j.
\]

Combining the sufficient and necessary conditions, we obtain the upper bound parameterized by the dataset diameter $c$:
\[
\operatorname{PLDiv}(\mathcal{X})
\le
\frac{n-1}{4}c^2.
\]

Moreover, equality holds if and only if the metric space is strictly equidistant:
\[
\begin{aligned}
& \operatorname{PLDiv}(\mathcal{X})
= \frac{n-1}{4}c^2\\
&\Longleftrightarrow
D(x_i,x_j)=c,\quad \forall i\neq j.
\end{aligned}
\]
\end{proof}

\item \textbf{Monotonicity.}
    Decreasing similarity increases diversity. For a fixed sample size, if pairwise similarity is decreased through a uniform dilation of the underlying metric $D$ to $\alpha D$, then every finite persistence interval \((b_i,d_i)\) becomes \((\alpha b_i,\alpha d_i)\) and
    \[
    \mathrm{PLDiv}(\alpha \mathcal{X})=\alpha^{2}\,\mathrm{PLDiv}(\mathcal{X}).
    \]
    Hence, when $\alpha>1$ and $\mathrm{PLDiv}(\mathcal{X})>0$, the score strictly increases; when $0<\alpha<1$, it strictly decreases.

    \begin{proof}
    Under the uniform dilation described above, each persistence lifetime scales as
    \[
    d_i-b_i \mapsto \alpha(d_i-b_i).
    \]
    By Proposition 4.2,
    \[
    \begin{aligned}
    \mathrm{PLDiv}(\alpha\mathcal{X})
    &=\frac14\sum_i(\alpha(d_i-b_i))^2\\
    &=\alpha^2\cdot\frac14\sum_i(d_i-b_i)^2\\
    &=\alpha^2\mathrm{PLDiv}(\mathcal{X}).
    \end{aligned}
    \]
    Hence, PLDiv strictly increases for $\alpha>1$ and strictly decreases for $0<\alpha<1$, whenever $\mathrm{PLDiv}(\mathcal{X})>0$, establishing the desired monotonicity under uniform changes in pairwise similarity.
\end{proof}

\item \textbf{Twin property.}
Adding an exact duplicate of a point does not change $\mathrm{PLDiv}(\mathcal{X})$.
    Let $\mathcal{X}$ be a dataset and let $x_i \in \mathcal{X}$. For the set $\mathcal{X}' = \mathcal{X} \cup \{x_n\}$ where $x_n = x_i$, the diversity is unchanged:
    \[
    \mathrm{PLDiv}(\mathcal{X}') = \mathrm{PLDiv}(\mathcal{X}).
    \]

\begin{proof}
A duplicate point at exactly the same coordinates is at zero distance from its twin. In the usual filtrations built from pairwise distances (e.g., Vietoris--Rips), the duplicate component is born at radius $0$ and immediately merges with its twin also at radius $0$. Hence the corresponding birth--death pair is $(0,0)$ and has lifetime $0$, contributing $(d-b)^2/4=0$ to the PLDiv sum. All other birth--death pairs are unchanged as well. Therefore $\mathrm{PLDiv}$ is unchanged.
\end{proof}

\item \textbf{Symmetry.}
PLDiv is invariant to the ordering of data points (permutation invariance).
    Since persistent homology depends only on the metric structure of $\mathcal{X}$ and $\mathrm{PLDiv}(\mathcal{X})$ is computed from the multiset of intervals $\{(b_i,d_i)\}$, relabeling or reordering points does not affect the value of the score.
    Let $\mathcal{X}=(x_1,\dots,x_n)$ be an ordered sequence of points and let $\pi$ be any permutation of $\{1,\dots,n\}$. For the permuted sequence $\mathcal{X}_\pi=(x_{\pi(1)},\dots,x_{\pi(n)})$, we have
    \[
    \mathrm{PLDiv}(\mathcal{X}_\pi) = \mathrm{PLDiv}(\mathcal{X}).
    \]

\begin{proof}
    The PH pipeline begins with the pairwise distance matrix $D$, where $D_{ij}=d(x_i,x_j)$. Let $\mathcal{X}_\pi$ be the reordered dataset. The distance matrix $D_\pi$ for the permuted data has entries $(D_\pi)_{ij}=d(x_{\pi(i)},x_{\pi(j)})$.
    Importantly, the set of all unique pairwise distances
    \[
    \{d(x_i,x_j)\}_{1 \leq i < j \leq n}
    \]
    is unchanged for both $\mathcal{X}$ and $\mathcal{X}_\pi$. The construction of the Vietoris--Rips filtration depends only on these distances. Hence, the persistence diagrams and lifetimes $\{l_i\}$ are identical. Therefore, any diversity measure computed from these lifetimes is invariant under permutation of the data and $\mathrm{PLDiv}$ is symmetry.

\end{proof}

\end{itemize}

\subsection{Axiomatic Properties Comparison}
\label{app:apc}

\begin{table}[h]
\centering
\setlength{\belowcaptionskip}{\baselineskip}
\caption{\textbf{Axiomatic comparison of PLDiv and baseline diversity metrics.} ``Partial'' indicates that the property holds only under additional kernel or normalization conditions. }
\label{tab:axiom-comparison}

\begin{tabular}{lcccc}
\toprule
\textbf{Property} & \textbf{PLDiv} & \textbf{MagArea} & \textbf{DCScore} & \textbf{Vendi} \\
\midrule
Effective size & Yes & Yes & Yes & Yes \\
Twin property & Yes & Yes & Yes & Partial \\
Monotonicity & Yes & Yes & Yes & No \\
Symmetry & Yes & Yes & Yes & Yes \\
\bottomrule
\end{tabular}
\end{table}

The foundational axioms for diversity measurement were introduced by \citet{leinster2012measuring} in the context of ecological diversity and later extended in related work on entropy and similarity-based diversity. These axioms include partitioning properties (e.g., effective number and replication), elementary properties (e.g., symmetry and identical species), and similarity-sensitive properties such as monotonicity. In our setting, the terminology is adapted to metric-space data: \textit{effective size} corresponds to the classical effective number property, while the \textit{twin property} is the metric-space analogue of the identical-species condition.

Table~\ref{tab:axiom-comparison} summarizes whether PLDiv and competing diversity measures satisfy several commonly studied axiomatic properties.
All methods satisfy effective size and symmetry, while PLDiv, MagArea, and DCScore additionally satisfy monotonicity and the twin property. Vendi Score only partially satisfies the twin property because this behavior depends on the choice of kernel and normalization, and it does not generally satisfy monotonicity. Instead, Vendi is more closely aligned with the partitioning and replication framework emphasized by \citet{leinster2012measuring}.

\section{Detailed Experiment Descriptions}

\subsection{Implementation of Gaussian Separation}
\label{app:gaussian}

\paragraph{Data simulations in Figure \ref{fig:pipeline}:} We draw two base samples $A,B\in\mathbb{R}^{50\times 2}$ independently from $\mathcal{N}(0,0.08I_2)$. For each separation value $s\in\{0,0.4,0.8,1.2,1.6,2.0\}$, we construct
\[
\begin{aligned}
X_s
&=\{a_i-(s/2,0):a_i\in A\}\\
&\quad \cup \{b_i+(s/2,0):b_i\in B\}.
\end{aligned}
\]
Reusing the same base samples across all $s$ keeps within-cluster noise fixed and makes inter-cluster separation $\mathrm{sep}$ the only changing factor.

\subsection{Implementation of PCA Intrinsic Dimension}
\label{tab:pca-dim}

We use a fixed simulated dataset with $n=3000$ points in an ambient dimension $d=50$. The full data matrix is generated as
\[
\begin{aligned}
X_{\mathrm{full}} &= ZD_v,\\
D_v &= \operatorname{diag}(\sqrt{v_1},\ldots,\sqrt{v_{50}}),\\
Z_{ij} &\sim \mathcal{N}(0,1),
\end{aligned}
\]
where $Z\in\mathbb{R}^{3000\times 50}$ is a standard Gaussian random matrix with independent entries $Z_{ij}\sim\mathcal{N}(0,1)$, and right-multiplication by $D_v$ rescales the $j$-th coordinate by $\sqrt{v_j}$. The resulting coordinate variances decrease linearly from $v_1=5.0$ to $v_{50}=0.1$.
This construction ensures that adding more PCA components introduces genuine lower-variance structure rather than pure noise. For each target intrinsic dimension $k\in\{2,4,8,12,16,24,32\}$, we fit standard centered PCA on the same $X_{\mathrm{full}}$, project it to $k$ components, and inverse-transform it back to the original ambient dimension:
\[
X^{(k)}
=\operatorname{PCA}^{-1}_k\!\left(\operatorname{PCA}_k(X_{\mathrm{full}})\right)
\in\mathbb{R}^{3000\times 50}.
\]
The reconstructed point cloud has rank $k$ up to numerical precision. All metric values are computed on $X^{(k)}$, so the sample size and ambient representation dimension remain fixed while the retained intrinsic dimensional complexity increases with $k$.

\begin{figure}[t]
    \centering
    \includegraphics[width=1.0\linewidth]{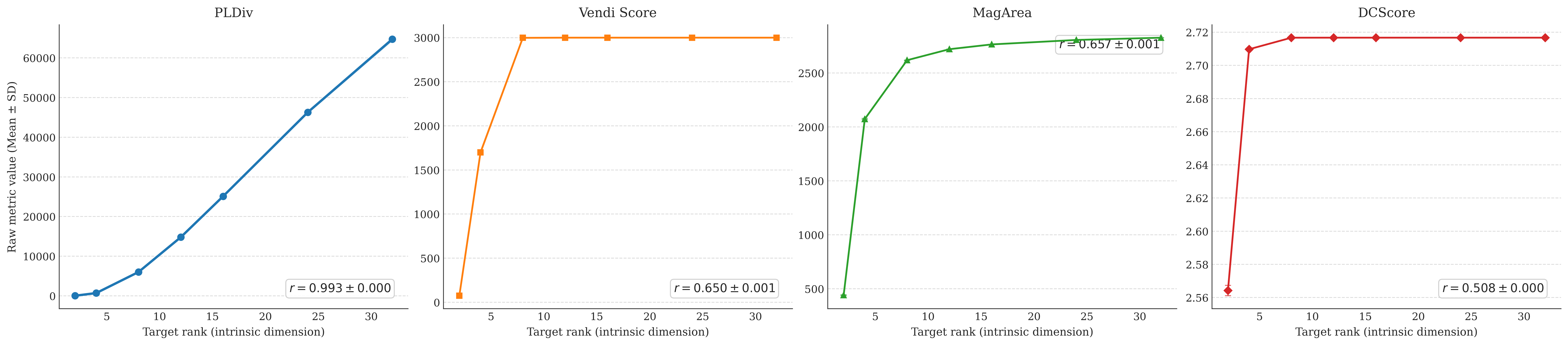}
    \caption{\textbf{PCA synthetic experiment (\textbf{raw}, unnormalized variance).} Each subplot shows the raw diversity score versus target intrinsic rank $k$ over 10 independent trials (error bars and shaded bands denote $\pm$1~SD). Text boxes report the mean Pearson correlation $r$.}
    \label{fig:pca-raw}
\end{figure}

\begin{figure}[!h]
    \centering
    \includegraphics[width=1.0\linewidth]{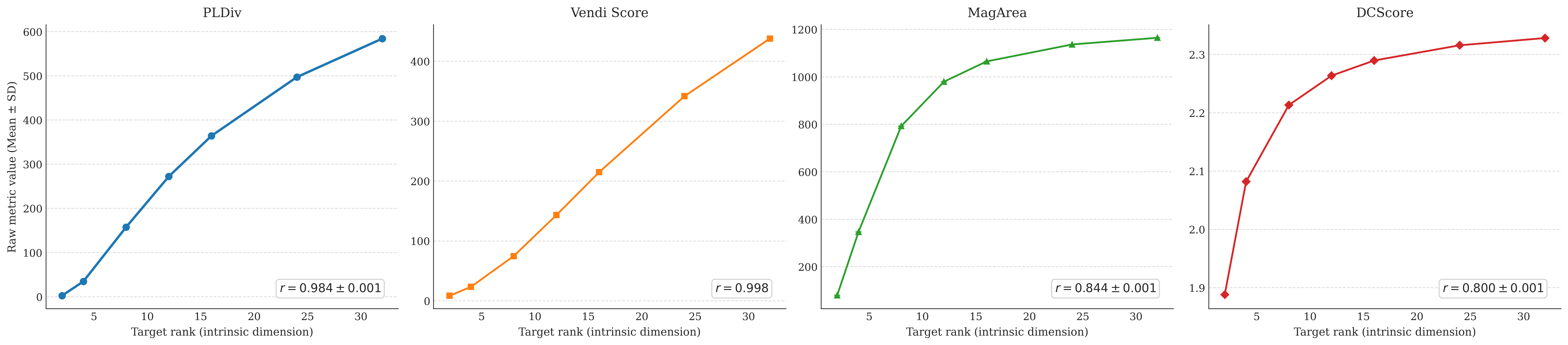}
    \caption{\textbf{PCA synthetic experiment (\textbf{trace-normalized}, fixed energy).} Same data and PCA reconstructions as Figure~\ref{fig:pca-raw}, but each point cloud is rescaled to unit covariance trace before metric evaluation, removing the scale confound.}
    \label{fig:pca-trace-norm}
\end{figure}

We consider two experimental conditions. In the \textbf{raw} setting (Figure~\ref{fig:pca-raw}), the reconstructions $X^{(k)}$ are used directly; since higher-rank reconstructions retain more variance, the total energy (covariance trace) grows with $k$, mirroring a realistic scenario where dimensionality and scale co-vary. In the \textbf{trace-normalized} setting (Figure~\ref{fig:pca-trace-norm}), we rescale each $X^{(k)}$ so that its sample covariance trace equals one, isolating the effect of intrinsic rank from overall scale.


Vendi Score and DCScore use the RBF kernel as the similarity kernel, and MagArea is set to the default.
In the raw setting (Figure~\ref{fig:pca-raw}), where total variance grows with $k$, PLDiv achieves the highest Pearson correlation ($r=0.993$) among all metrics, while Vendi Score ($r=0.650$) and DCScore ($r=0.508$) exhibit early saturation due to their sensitivity to the co-varying scale, and MagArea ($r=0.657$) is similarly affected. When the scale confound is removed via trace normalization (Figure~\ref{fig:pca-trace-norm}), all kernel-based metrics improve substantially: Vendi Score rises to $r=0.998$, MagArea to $r=0.844$, and DCScore to $r=0.800$. This suggests that their lower correlations in the raw condition may be largely attributed to the entangled scale, rather than a fundamental inability to track dimensionality.
Critically, PLDiv remains consistently strong in both settings ($r=0.984$ under normalization), showing that PLDiv robustly tracks intrinsic dimension across both the raw and trace-normalized settings. This robustness is a practical advantage: in real-world datasets, scale and structural complexity often co-vary, and a metric that conflates the two may over- or under-estimate diversity.

\subsection{Imbalanced Long-tail data}
\label{appendix:tail}
\begin{figure}[h!]
    \centering
    \includegraphics[width=1.0\linewidth]{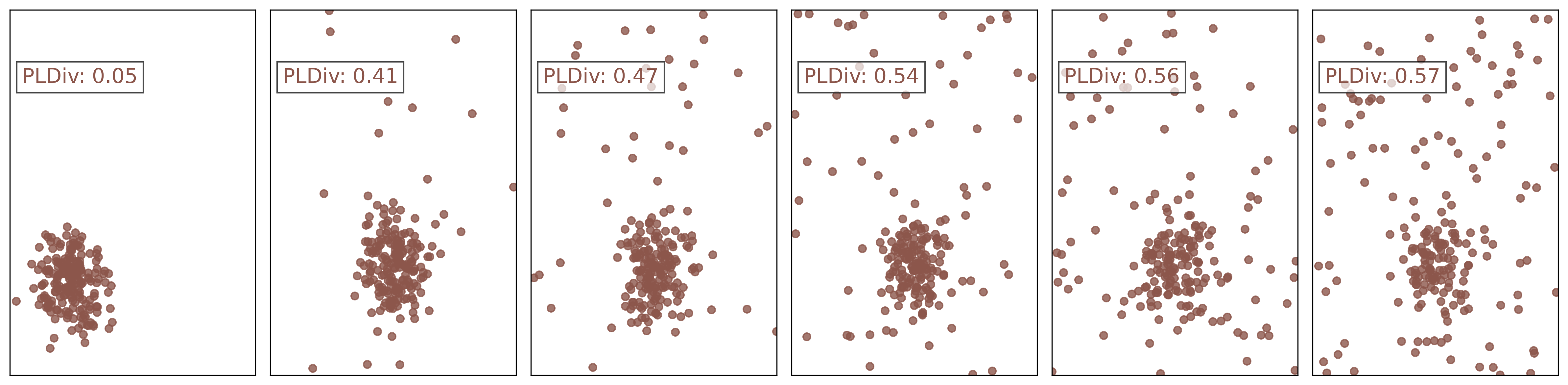}
    \caption{\textbf{PLDiv can reliably predict diversity in imbalanced data, where diversity increases monotonically.}}
    \label{fig:tail}
\end{figure}

To explore how PLDiv performs on imbalanced data, we generated a series of small long-tail datasets. First, we utilized D4 in synthetic toy examples, which form a single cluster with 200 data points. To simulate long-tail effects, outlier points were added uniformly within a square region in varying amounts of 20, 40, 60, 80, and 100 samples, while keeping the cluster size at 200 - $n_{\text{outliers}}$. Each variant thus exhibits increasing imbalance between the dense Gaussian core and sparse tail regions. Figure \ref{fig:tail} demonstrates that PLDiv effectively handles the imbalanced dataset.

\subsection{Implementation of Text Embeddings}
\label{appedix:text}
We evaluate PLDiv as a metric of semantic diversity using the dataset from \citet{tevet-berant-2021-evaluating}, comprising 1,000 input contexts from three tasks. Ten outputs per input context were generated by varying the softmax temperature ($dec$), and a subset of 200 input contexts was human-annotated to obtain mean diversity scores (\textit{ABS-HDS}).
Text embedding models we used are listed below:

\begin{itemize}
    \item all-MiniLM-L12-v2: general text embedding model, dimension 384
    \item all-mpnet-base-v2: general text embedding model, dimension 768
    \item bert-large-nli-stsb-mean-tokens:  general text embedding model, dimension 1024
    \item Qwen3-Embedding-4B: advanced LLM-based embedding models, dimension 2560
    \item Qwen3-Embedding-8B: advanced LLM-based embedding models, dimension 4096
\end{itemize}

\begin{figure}[ht]
    \centering
    \includegraphics[width=1\columnwidth]{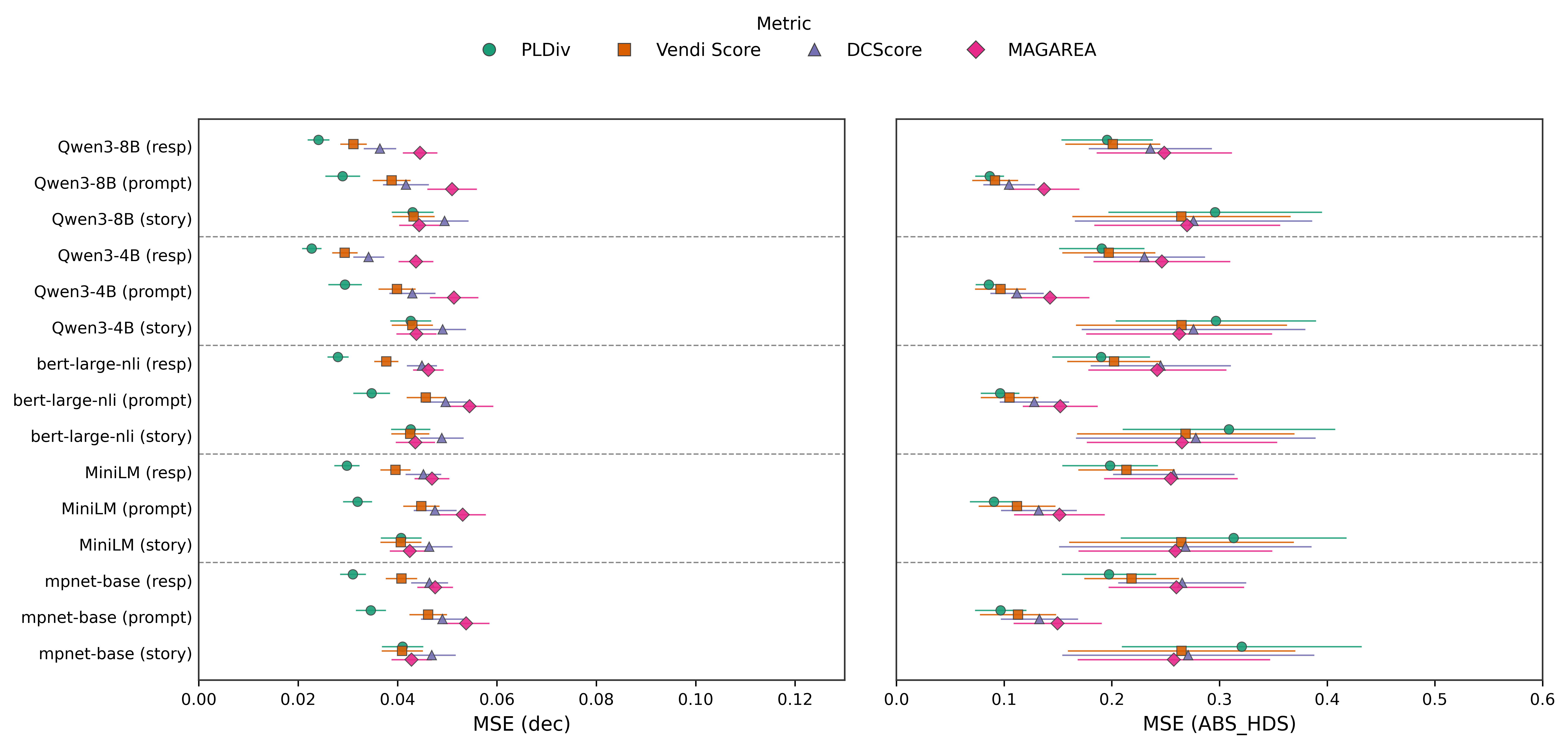}
    \caption{\textbf{MSE for four metrics evaluated against temperature $dec$ and human diversity score \textit{ABS-HDS} under the Euclidean/RBF setting.} Error bars show the standard deviation across repeated five-fold cross-validation scores.}
    \label{fig:text_mse}
\end{figure}

Figure \ref{fig:text_mse} reports the mean squared error (MSE) of the linear regressions for softmax temperature $dec$ and mean human-annotated diversity score (\textit{ABS-HDS}) under the Euclidean/RBF setting. The MSE results reinforce the task-dependent pattern observed with the correlation-based measures: PLDiv consistently performs well for prompt completion and response generation under both evaluation targets, while the human-evaluated story task favors other metrics. This consistency across evaluation measures further supports the robustness of the main findings.

\begin{figure}[h!]
    \centering

    \includegraphics[width=0.55\columnwidth]{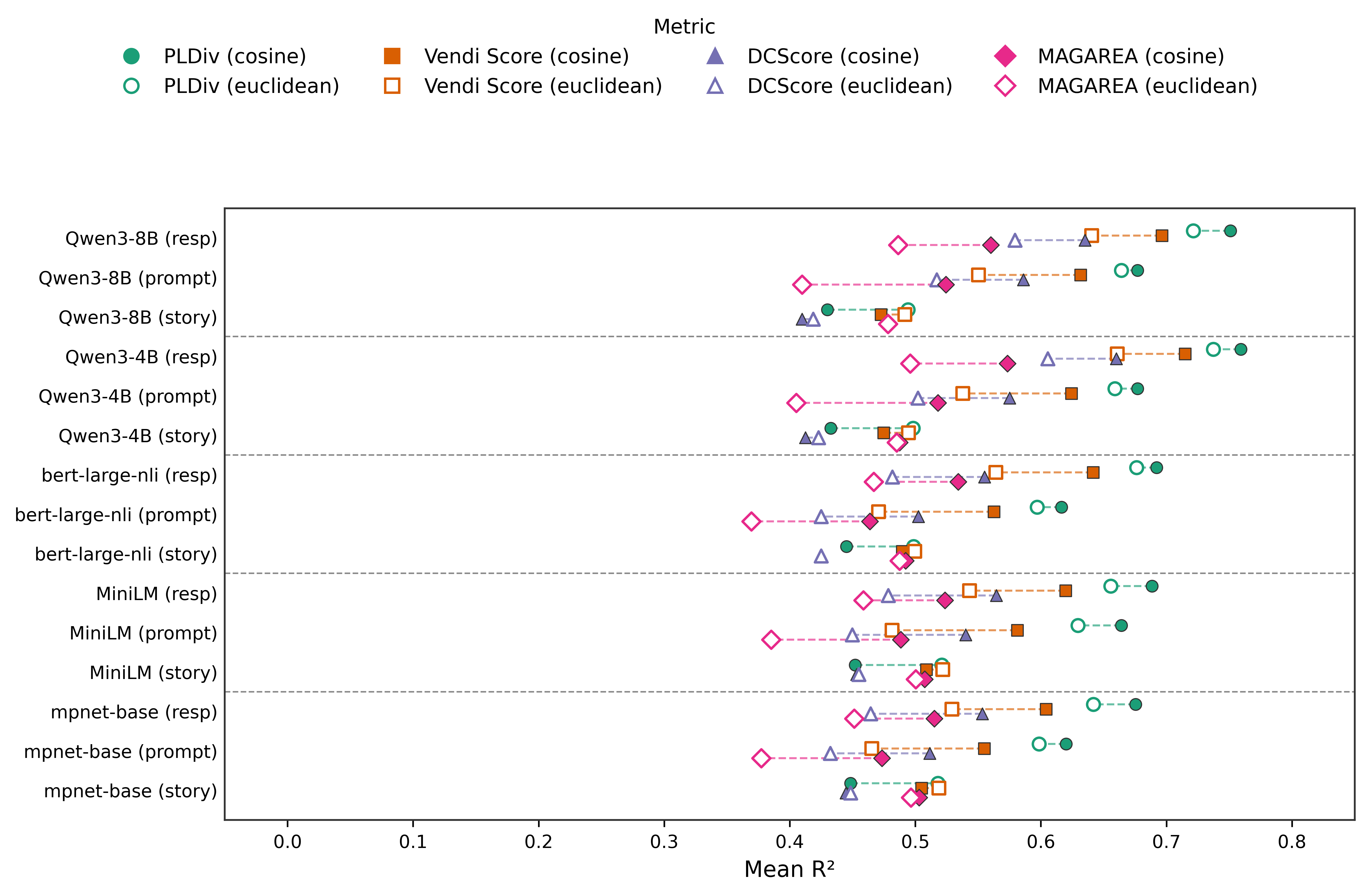}

    \vspace{2mm}

    \includegraphics[width=0.55\columnwidth]{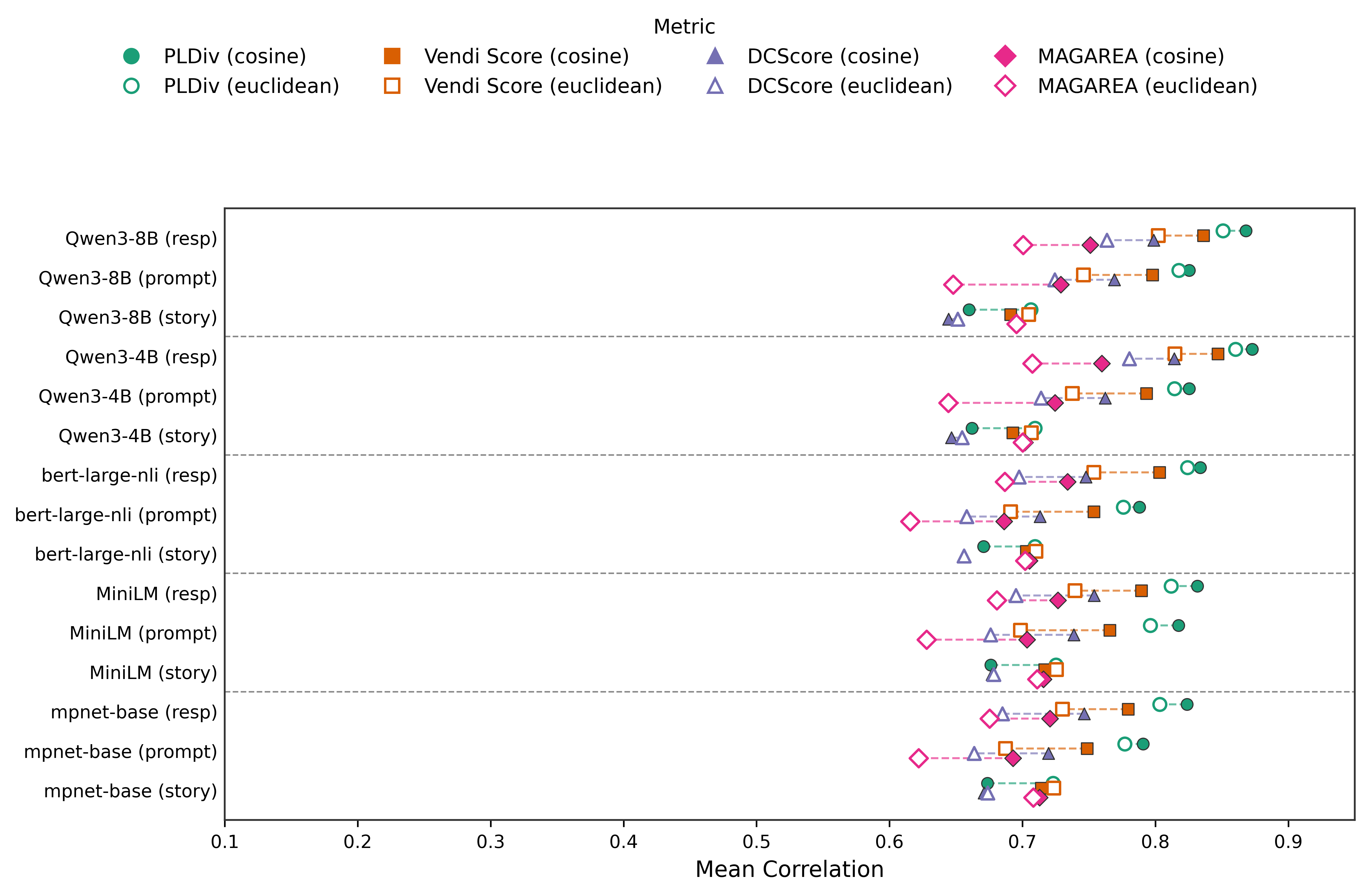}

    \vspace{2mm}

    \includegraphics[width=0.55\columnwidth]{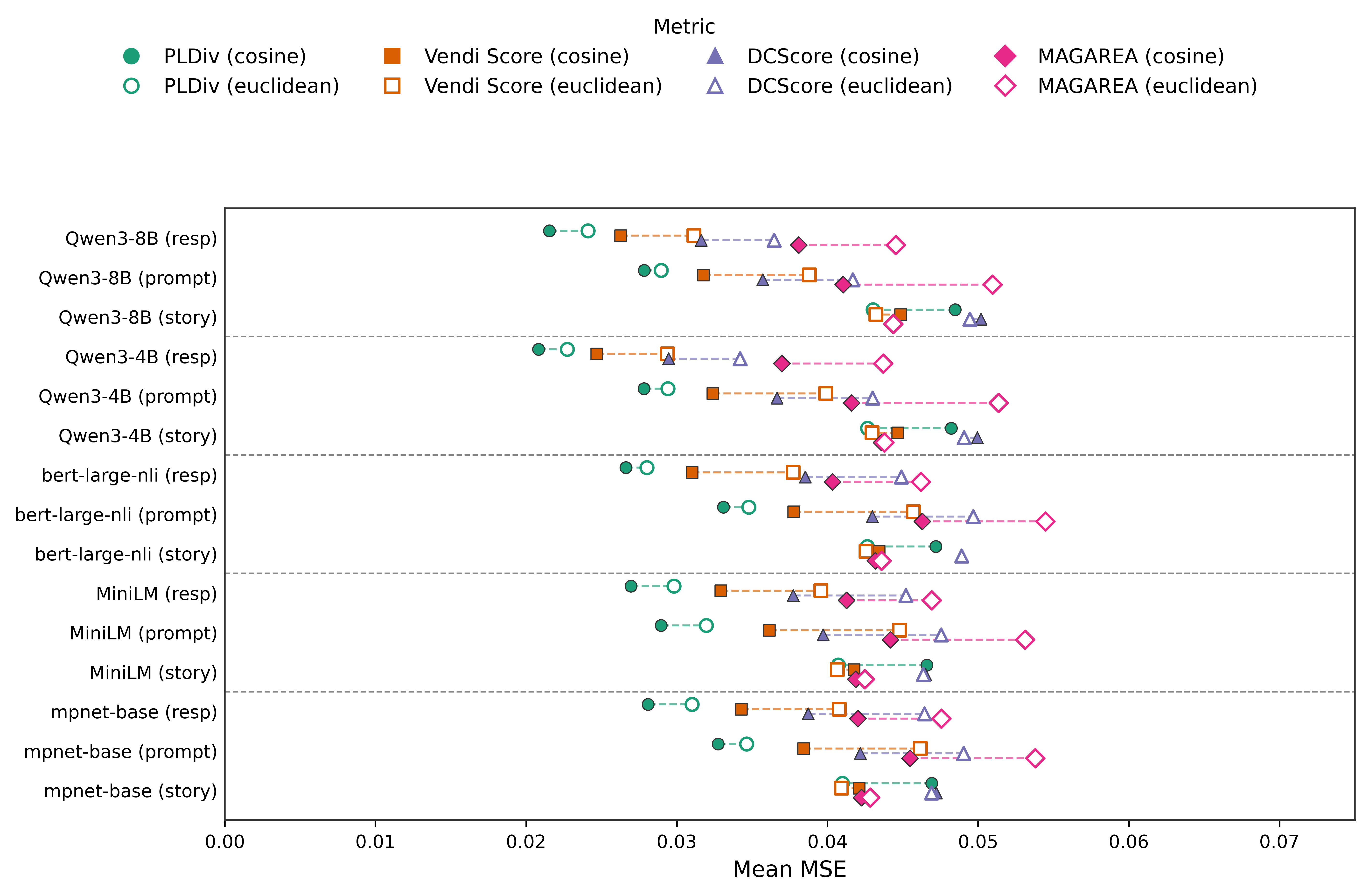}

    \caption{\textbf{Diversity metric performance is evaluated across different distance/similarity matrices for the temperature tasks.} Filled markers denote cosine distance/similarity and open markers denote Euclidean distance or, for Vendi Score and DCScore, the RBF kernel. PLDiv consistently and reliably outperforms other metrics across various embedding models and distance matrices.}
    \label{fig:cs_eu}
\end{figure}

To explore the impact of the distance/similarity matrix, we applied both cosine distance/similarity and Euclidean distance/RBF kernel as inputs in this experiment for the temperature (dec) tasks. Figure \ref{fig:cs_eu} demonstrates that PLDiv consistently and reliably outperforms other metrics across various embedding models and distance matrices.

\subsubsection{Duplicate Diagnostics for Story Completion under \textit{ABS-HDS} Evaluation}
\label{app:text-duplicate-sensitivity}

An exact duplicate is a generated response whose string occurs at least twice within the same ten-response set. Because the embedding function is deterministic, exact text duplicates also produce identical embeddings. Exact duplication is substantially more common in the human-evaluated story task: 118 of 203 response sets contain a duplicate, compared with 35 of 203 prompt sets and 53 of 203 response-generation sets. Because exact duplicates directly reduce the number of unique responses in a set, we examine whether the human diversity score \textit{ABS-HDS} is associated with unique-response count. The association is strongest for story completion ($r=0.817$), compared with prompt completion ($r=0.682$) and response generation ($r=0.622$). Together with the higher prevalence of exact duplicates in story completion, this result indicates that response repetition is more closely associated with human diversity scores in the story task.

This distinction affects the metrics differently. Vendi Score and DCScore retain sample multiplicity through the kernel spectrum and softmax separability, respectively. In contrast, exact duplicates do not contribute additional positive persistence to PLDiv. With the number of generated responses held fixed, replacing a distinct response with a duplicate can still reduce PLDiv by removing distinct semantic information, but the repeated copy itself does not introduce an additional diversity contribution.

As a post-hoc sensitivity analysis, we keep only the 85 story response sets in which all ten generated responses are unique. Table~\ref{tab:story-all-unique} reports Pearson correlations with \textit{ABS-HDS} under the Euclidean/RBF setting. PLDiv has the highest correlation for all five encoders after the exact duplicate instances are removed.

\begin{table}[H]
    \centering
    \small
    \setlength{\abovecaptionskip}{\baselineskip}
    \setlength{\belowcaptionskip}{\baselineskip}
    \caption{\textbf{Post-hoc decuplicate story task for Pearson correlations with \textit{ABS-HDS} ($n=85$ response sets with ten unique outputs).}}
    \label{tab:story-all-unique}

    \begin{tabular}{lcccc}
        \toprule
        Embedding Model & PLDiv & Vendi & DCScore & MagArea \\
        \midrule
        Qwen3-4B       & \textbf{0.626} & 0.593 & 0.572 & 0.560 \\
        Qwen3-8B       & \textbf{0.613} & 0.585 & 0.567 & 0.548 \\
        MiniLM-L12-v2  & \textbf{0.590} & 0.562 & 0.556 & 0.550 \\
        MPNet-base-v2  & \textbf{0.549} & 0.541 & 0.528 & 0.537 \\
        BERT-large-NLI & \textbf{0.590} & 0.572 & 0.550 & 0.557 \\
        \bottomrule
    \end{tabular}
\end{table}

\subsection{Implementation of Instruction Near-Duplicate Collapse}
\label{app:instruction-collapse}

The instruction-data experiment in Section~\ref{sec:instruction-collapse} is designed to isolate local semantic redundancy from category imbalance. Let $S_0$ be a balanced collection of $N=400$ instructions sampled from Dolly-15K categories. At collapse level $\rho$, we retain $m=(1-\rho)N$ anchors $A_\rho\subset S_0$ and construct $M=\rho N$ replacements. Let $(a_1,\ldots,a_M)$ be anchors sampled with replacement from $A_\rho$ according to the category-specific replacement counts. For each $a_r$, the replacement is the nearest previously unused instruction outside $S_0$ from the same category:
\[
x_r^\star
=
\operatorname*{arg\,max}_{\substack{
x\notin S_0\cup\{x_1^\star,\ldots,x_{r-1}^\star\}\\
c(x)=c(a_r)
}}
\operatorname{sim}\bigl(f(x),f(a_r)\bigr),
\]
where $f(\cdot)$ is the sentence embedding function and $\operatorname{sim}(\cdot,\cdot)$ denotes cosine similarity. Let $R_\rho=(x_1^\star,\ldots,x_M^\star)$ denote the resulting replacements. The final dataset $S_\rho$ consists of the retained anchors $A_\rho$ and replacements $R_\rho$, with $|S_\rho|=N$.

If a category does not contain enough eligible external neighbors, the remaining replacements are sampled with replacement from the retained anchors in that category. This construction keeps the dataset size fixed and the category mixture approximately fixed while progressively increasing local semantic redundancy.

The experiment uses $\rho\in\{0,0.1,0.25,0.5,0.75\allowbreak\}$ and ten random repeats. The validation target is the non-redundant fraction $1-\rho$. In the produced datasets, the mean cosine similarity between anchors and their nearest-neighbor replacements is around 0.5, while random same-category pairs have cosine similarity around 0.1, confirming that the manipulation creates near-duplicate collapse rather than arbitrary replacement.

\begin{table}[h!]
\centering
\setlength{\abovecaptionskip}{\baselineskip}
\setlength{\belowcaptionskip}{\baselineskip}
\caption{\textbf{Statistics of the near-duplicate collapse benchmark across five collapse levels $\rho$, aggregated over 10 trials (reported as mean $\pm$ standard deviation).} The \textit{Anchors} and \textit{Near dups.} columns denote the number of retained anchor examples and nearest-neighbor replacement examples, respectively. \textit{Anchor NN cos.} reports the mean cosine similarity between anchors and their nearest-neighbor replacements, while \textit{Random cat. cos.} reports the cosine similarity between anchors and randomly selected examples from the same task category. The substantially higher similarity of nearest-neighbor pairs confirms that the construction procedure introduces controlled local redundancy while preserving dataset size and category composition.}
\label{tab:instruction-collapse-diagnostics}

\setlength{\tabcolsep}{3pt}
\begin{tabular}{lccccc}
\toprule
$\rho$ & Anchors & Near dups. & PLDiv & Anchor NN cos. & Random cat. cos. \\
\midrule
0.00 & 400 & 0   & $115.51 \pm 1.24$ & --                & --                \\
0.10 & 360 & 40  & $112.22 \pm 1.21$ & $0.53 \pm 0.03$ & $0.11 \pm 0.02$ \\
0.25 & 300 & 100 & $107.93 \pm 1.06$ & $0.51 \pm 0.01$ & $0.12 \pm 0.01$ \\
0.50 & 200 & 200 & $99.72 \pm 2.06$  & $0.52 \pm 0.01$ & $0.11 \pm 0.01$ \\
0.75 & 100 & 300 & $92.72 \pm 1.41$  & $0.49 \pm 0.01$ & $0.11 \pm 0.00$ \\
\bottomrule
\end{tabular}
\end{table}

\subsection{Implementation of Reasoning Homogenization}
\label{app:h1-reasoning-homogenization}

These implementation details describe the experiment introduced in Section \ref{sec:h1-reasoning-homogenization}. The reasoning-path homogenization is for the task where the same problem, a model or data pipeline may preserve multiple answer traces while gradually making their intermediate reasoning states equivalent to one consensus trajectory.

\paragraph{Data.}
We use ProcessBench \citep{zheng2024processbench}, which contains \texttt{gsm8k}, \texttt{math}, \texttt{olympiadbench}, and \texttt{omnimath} splits from HuggingFace. We then filter the data to require that all selected paths have correct final answers, contain at least two nonempty steps per path, and come from distinct generators. The prototype path is chosen from traces with all intermediate steps correct when available; otherwise, it is chosen from traces with correct final answers. The branch path is another trace with the correct final answer from the same problem. The final set contains 90 problems (88 \texttt{olympiadbench}, 2 \texttt{gsm8k}). Every trial uses all 90 problems and varies only the random replacement order at intermediate states with homogenization levels $p$.

\paragraph{Collapse Construction.}
To preserve the temporal nature of the reasoning trajectory, we use a graph to represent the reasoning states. For each selected problem, let $A=(a_1,\ldots,a_m)$ be the prototype path and $B=(b_1,\ldots,b_n)$ be the alternative branch. The graph contains a shared question node $q$, a shared answer node $z$, and nodes for all steps in both paths:
\[
\begin{aligned}
q &\rightarrow a_1 \rightarrow \cdots \rightarrow a_m \rightarrow z,\\
q &\rightarrow b_1 \rightarrow \cdots \rightarrow b_n \rightarrow z .
\end{aligned}
\]
All edges along the paths have weight 1. Since the raw reasoning paths may have different lengths, each branch step $b_j$ is aligned to the prototype step $a_{\alpha(j)}$ using relative position:
\[
\alpha(j) = 1 + \operatorname{round}\!\left(\frac{j-1}{n-1}(m-1)\right),
\]
with special handling for the case when the path length is one. For each trial, the $n$ branch steps are randomly permuted to determine the replacement order. According to the homogenization level $p\in\{0,0.25,0.5,0.75,1.0\}$, the first $pn$ steps in this ordering have their texts replaced, while the graph structure and temporal order remain unchanged.

We use \texttt{gpt-5.6} to generate replacement texts for the alternative branch steps. Each candidate replacement must satisfy three criteria: (i) it must differ from both the prototype and the original branch text; (ii) it must preserve the prototype's canonical number set exactly; and (iii) it must pass a second \texttt{gpt-5.6} audit verifying consistency in reasoning state, mathematical operation, content, scope, and single-step conciseness. For every replaceable branch step and every homogenization level $p$, there is an aligned cross-edge. Let $t_j^{(p)}$ denote the effective text of branch node $b_j$ at level $p$, and let $\phi$ be the normalized embedding function. The semantic edge weight is defined as
\[
\begin{aligned}
w_j^{(p)} &= \max\!\bigl(10^{-6},\, 1-\cos\!\left(\phi(t_j^{(p)}),\phi(a_{\alpha(j)})\right)\bigr).
\end{aligned}
\]
The cross-edge set remains the same across all intervention levels, and the path-chain edges keep a unit weight. Therefore, the homogenization level $p$ affects only the semantic edge weights, not the combinatorial topology of the graph.

We embed the effective reasoning-step texts with \texttt{all-MiniLM-L6-v2}, use the semantic cross-edge weights to construct the weighted trajectory graph, and compute one normalized distance matrix $D_p$ for each problem, trial, and intervention level. All structural metrics receive this same $D_p$. PLDiv computes $H_0$ and $H_1$ Rips persistence from $D_p$. Vendi Score and DCScore receive the same graph kernel $K_p=\exp(-D_p^2/(2\sigma^2))$ with $\sigma=1$, and MagArea is integrated from the same $D_p$ over 100 scales on $[0,10]$. Distinct-1 and Distinct-2 are computed on the effective reasoning-step texts. Inverse Self-BLEU is one minus mean leave-one-trace-out smoothed BLEU-4, treating each complete reasoning path as one sample and using a math-aware tokenizer that retains \LaTeX{} commands, numbers, identifiers, operators, and punctuation.

\paragraph{Evaluation.}
We quantify each metric's sensitivity to reasoning collapse by comparing its average values at the endpoints of no intervention $p=0$ and full intervention $p=1$ for each problem $i$:
\[
\Delta_i^{(m)}
=
\overline{M}_{i,1}^{(m)}
-
\overline{M}_{i,0}^{(m)},
\]
where $\overline{M}_{i,p}^{(m)}$ is the value of metric $m$ for problem $i$ at intervention level $p$, averaged over replacement-order trials. We first average the endpoint changes across the $N$ problems:
\[
\overline{\Delta}^{(m)}
=
\frac{1}{N}\sum_{i=1}^{N}\Delta_i^{(m)}.
\]
Concord. counts how many individual problems move in the same direction as this overall change:
\[
\operatorname{Concord}^{(m)}
=
\begin{cases}
\#\{i:\Delta_i^{(m)}<0\}, & \text{if }\overline{\Delta}^{(m)}<0,\\
\#\{i:\Delta_i^{(m)}>0\}, & \text{if }\overline{\Delta}^{(m)}>0.
\end{cases}
\]
The table reports this count out of $N=90$. In words, if the metric decreases overall, Concord. is the number of problems for which it decreases; if the metric increases overall, it is the number for which it increases. It is not a correlation or a summary of the plotted error bands.

For full-range prediction, metric $m$ is converted to the problem-relative contraction feature
\[
x_{i,p,t}^{(m)}=\log\!\left(\frac{M_{i,0,t}^{(m)}+10^{-12}}{M_{i,p,t}^{(m)}+10^{-12}}\right)
\]
A separate ordinary least-squares model with an intercept is fit for each metric and replacement-order trial to predict the realized replacement fraction. 10-fold cross-validation is grouped by complete problem statement, and all five intervention levels from a held-out problem remain in the same fold. We concatenate the out-of-fold predictions before computing $R^2$, and RMSE, and then report the mean and sample standard deviation of these scores across the 10 trials.

\subsection{Implementation of Image Embeddings}

Section~5.4 uses a colorized version of MNIST \citep{deng2012mnist} and follows the class-count proxy studied by \citet{ospanov2024towards}. The experiment fixes the total sample size at $N=750$ for every condition and evaluates class counts $k\in\{1,\ldots,10\}$. In each of 10 repeats, we draw a random permutation of the 10 digit classes (seed 17); the condition with class count $k$ uses the first $k$ classes in that permutation, with the 750 samples allocated as evenly as possible across the selected classes. Thus, class count changes while sample size remains fixed. We compare PLDiv with Vendi Score, MagArea, and DCScore using cosine distance for PLDiv and MagArea and cosine similarity for Vendi Score and DCScore. Inception V3 and ResNet-18 provide 2048- and 512-dimensional embeddings, respectively. Figure~\ref{fig:image} reports the mean curves and variability across the 10 randomized class orders.

\subsection{Scalable Sparse PLDiv}
\label{app:knn-sparse-pldiv}

For $H_0$ persistent homology under the Vietoris--Rips filtration, the non-infinite death times are exactly the edge weights of the minimum spanning tree of the complete weighted graph on the dataset. Therefore,
\[
\begin{aligned}
\mathrm{PLDiv}(\mathcal{X})
&=\frac{1}{4}\sum_{e\in \mathrm{MST}(K_n)} w(e)^2,
\end{aligned}
\]
where $K_n$ is the complete graph with edge weight $w(i,j)=d(x_i,x_j)$.

The scalable approximation replaces $K_n$ with a symmetrized $k$-nearest-neighbor graph. Let
\[
\begin{aligned}
E_k
&=\{(i,j): j\in \mathrm{kNN}(i)\\
&\qquad\quad \text{or}\ i\in \mathrm{kNN}(j)\},\\
G_k&=(V,E_k,w).
\end{aligned}
\]
We compute
\[
\begin{aligned}
\widehat{\mathrm{PLDiv}}_k(\mathcal{X})
&=\frac{1}{4}\sum_{e\in \mathrm{MST}(G_k)} w(e)^2.
\end{aligned}
\]
If $G_k$ is connected and contains all edges of $\mathrm{MST}(K_n)$, then $\widehat{\mathrm{PLDiv}}_k=\mathrm{PLDiv}$.

\begin{algorithm}[ht]
\caption{kNN Sparse-MST PLDiv for $H_0$}
\begin{algorithmic}[1]
\REQUIRE Data $X=\{x_i\}_{i=1}^n$, metric $d$, neighborhood size $k$
\ENSURE Sparse PLDiv estimate $\widehat{\mathrm{PLDiv}}_k(X)$ and diagnostic flags
\STATE Build directed $k$NN edge set $E_k^{\rightarrow}$
\STATE Symmetrize edges to get $E_k$
\STATE Assign edge weights $w(i,j)=d(x_i,x_j)$ and form graph $G_k = (V, E_k, w)$
\IF{$G_k$ is disconnected}
    \STATE Record connected components number
    \STATE Increase $k$ or compute PLDiv componentwise as approximation
\ENDIF
\STATE Compute MST $T_k = \mathrm{MST}(G_k)$
\STATE \textbf{return} $\widehat{\mathrm{PLDiv}}_k(X) = \frac{1}{4} \sum_{e \in T_k} w(e)^2$
\STATE For validation, repeat with larger $k$ and compare with dense MST if feasible
\end{algorithmic}
\end{algorithm}

\paragraph{Implementation.}
In our implementation, a $k$-nearest-neighbor ($k$NN) graph is constructed using pairwise distance. The directed $k$NN graph is then symmetrized by taking the union of neighbor edges before computing the minimum spanning tree (MST) on the resulting sparse graph. We use $k=20$ as the default setting in large-scale experiments. The framework supports both exact nearest-neighbor search and approximate nearest-neighbor methods such as NN-Descent, HNSW, and FAISS for efficient scaling to larger datasets.

To evaluate the reliability of the sparse $k$NN approximation, we vary the neighborhood size $k$ and examine whether PLDiv remains stable across different graph constructions. The sparse graph serves as an approximation to the complete graph used in exact $H_0$ persistence computation, and increasing stability with larger $k$ indicates that the essential connectivity structure of the data is preserved.

\begin{figure}[h!]
    \centering
    \includegraphics[width=1\linewidth]{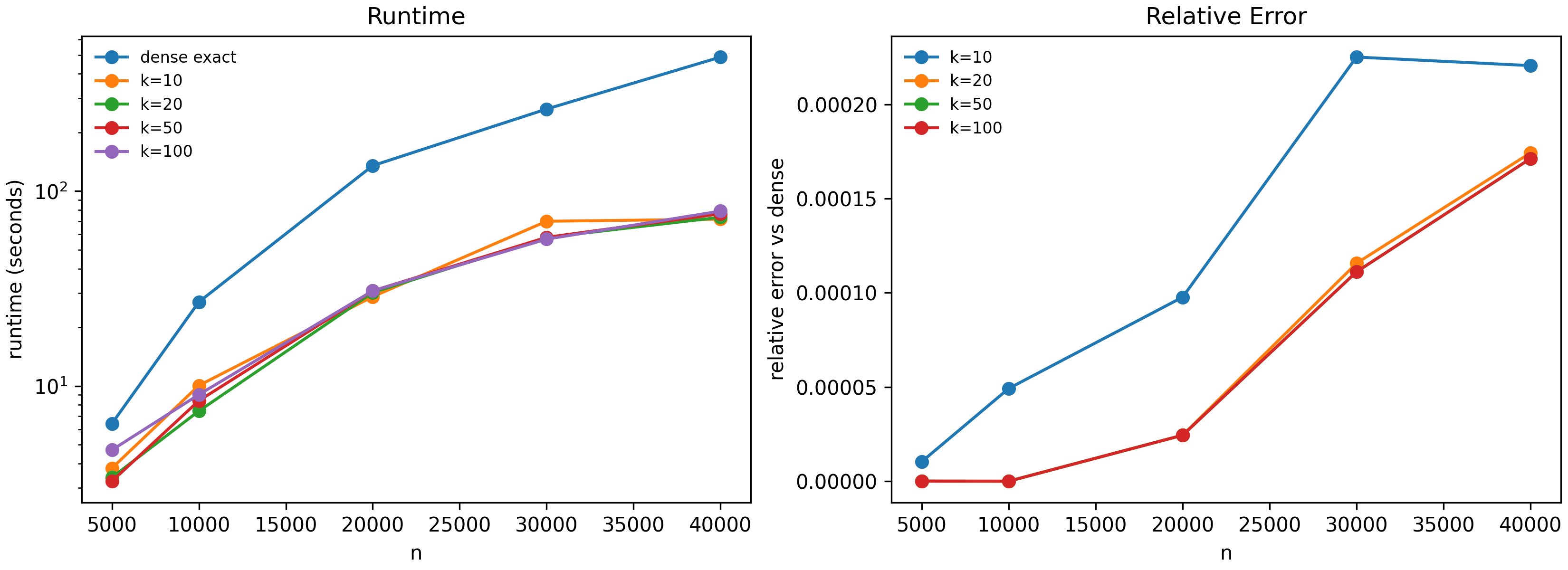}
    \caption{\textbf{kNN sparse-MST PLDiv runtime and validation error.} Dense exact PLDiv scales quadratically due to the full distance matrix; kNN sparse PLDiv remains fast and has low relative error on feasible exact-validation sizes.}
    \label{fig:knn-sparse-runtime}
\end{figure}

In our large-scale image-embedding experiments, PLDiv values remain nearly unchanged for $k\in{10,20,50,100}$ across sample sizes ranging from 5k to 40k, suggesting that $k=20$ provides a sufficient approximation for the reported experiments (Figure~\ref{fig:knn-sparse-runtime}, Table~\ref{tab:knn-sparse-k-ablation}).

\begin{table}[h]
\centering
\setlength{\abovecaptionskip}{\baselineskip}
\setlength{\belowcaptionskip}{\baselineskip}
\caption{\textbf{Sparse PLDiv values across different $k$ values in the kNN sparse-MST implementation on image embeddings with sample sizes from 5k to 40k.} The values are stable across $k$, indicating that small neighborhood graphs already preserve the relevant $H_0$ connectivity structure in this large-scale setting.}
\label{tab:knn-sparse-k-ablation}

\begin{tabular}{lcccc}
\toprule
\textbf{Sample size} & \textbf{$k=10$} & \textbf{$k=20$} & \textbf{$k=50$} & \textbf{$k=100$} \\
\midrule
5k  & 46.506 & 46.506 & 46.506 & 46.506 \\
10k & 78.014 & 78.010 & 78.010 & 78.010 \\
20k & 133.558 & 133.549 & 133.549 & 133.549 \\
30k & 184.946 & 184.926 & 184.925 & 184.925 \\
40k & 232.890 & 232.879 & 232.879 & 232.879 \\
\bottomrule
\end{tabular}
\end{table}

\paragraph{Complexity.}
Dense exact PLDiv requires $\mathcal{O}(n^2d)$ time to form all pairwise distances from $d$-dimensional embeddings, $\mathcal{O}(n^2)$ memory, and an MST computation over $\mathcal{O}(n^2)$ edges. After the sparse graph has been constructed, kNN sparse PLDiv stores $\mathcal{O}(kn)$ edges and computes the MST in $\mathcal{O}(kn\log n)$ time (Table \ref{tab:fair-runtime-three-metrics}). Exact brute-force kNN construction can still require pairwise search, but the downstream PLDiv computation no longer uses the dense complete graph; approximate nearest-neighbor construction can further reduce the front-end search cost for very large datasets.

\begin{table*}[t]
\centering
\setlength{\abovecaptionskip}{\baselineskip}
\setlength{\belowcaptionskip}{\baselineskip}
\caption{\textbf{Runtime comparison of diversity metrics on image embeddings.} The cosine setting uses cosine distance/similarity, with Vendi computed by its official dual formulation. The Euclidean/RBF setting uses normalized Euclidean distance for Sparse PLDiv and MagArea, and RBF kernels for Vendi and DCScore. Runtime (seconds) is reported as mean $\pm$ standard deviation over five runs; -- indicates settings not completed due to prohibitive runtime.}
\label{tab:fair-runtime-three-metrics}

\begin{tabular}{lccccc}
\toprule
\textbf{Metric} & \textbf{5k} & \textbf{10k} & \textbf{20k} & \textbf{30k} & \textbf{40k} \\
\midrule
Sparse PLDiv (cosine, $k=20$)
& $2.65_{\scriptscriptstyle \pm 0.54}$
& $6.68_{\scriptscriptstyle \pm 0.54}$
& $21.08_{\scriptscriptstyle \pm 1.33}$
& $41.83_{\scriptscriptstyle \pm 2.11}$
& $68.20_{\scriptscriptstyle \pm 2.37}$ \\
Vendi (cosine)
& $0.53_{\scriptscriptstyle \pm 0.11}$
& $0.76_{\scriptscriptstyle \pm 0.34}$
& $1.09_{\scriptscriptstyle \pm 1.09}$
& $1.58_{\scriptscriptstyle \pm 0.84}$
& $1.68_{\scriptscriptstyle \pm 0.96}$ \\
DCScore (cosine)
& $0.43_{\scriptscriptstyle \pm 0.10}$
& $0.95_{\scriptscriptstyle \pm 0.05}$
& $3.14_{\scriptscriptstyle \pm 0.07}$
& $6.44_{\scriptscriptstyle \pm 0.18}$
& $11.25_{\scriptscriptstyle \pm 0.39}$ \\
MagArea (cosine)
& $107.76_{\scriptscriptstyle \pm 1.16}$
& $425.06_{\scriptscriptstyle \pm 3.03}$
& -- & -- & -- \\
\midrule
Sparse PLDiv (Euclidean, $k=20$)
& $1.22_{\scriptscriptstyle \pm 0.06}$
& $2.82_{\scriptscriptstyle \pm 0.15}$
& $6.97_{\scriptscriptstyle \pm 0.18}$
& $12.48_{\scriptscriptstyle \pm 0.17}$
& $9.85_{\scriptscriptstyle \pm 0.20}$ \\
Vendi (RBF)
& $4.21_{\scriptscriptstyle \pm 2.78}$
& $17.69_{\scriptscriptstyle \pm 2.85}$
& $204.02_{\scriptscriptstyle \pm 6.66}$
& $808.32_{\scriptscriptstyle \pm 67.12}$
& $2083.00_{\scriptscriptstyle \pm 69.27}$ \\
DCScore (RBF)
& $0.72_{\scriptscriptstyle \pm 0.15}$
& $3.28_{\scriptscriptstyle \pm 0.30}$
& $9.09_{\scriptscriptstyle \pm 0.39}$
& $15.17_{\scriptscriptstyle \pm 0.32}$
& $24.78_{\scriptscriptstyle \pm 1.22}$ \\
MagArea (Euclidean)
& $103.75_{\scriptscriptstyle \pm 7.77}$
& $388.91_{\scriptscriptstyle \pm 5.82}$
& -- & -- & -- \\
\bottomrule
\end{tabular}
\end{table*}

PLDiv is designed to provide a principled measure of data diversity that complements existing diversity metrics. Our results demonstrate the practicality of Sparse PLDiv at scale, supporting its use for diversity measurement on large datasets.

\section{Computational Environment}

All experiments were conducted on a high-performance computing server equipped with an AMD EPYC 7413 24-Core Processor and an NVIDIA A100-80GB GPU. The software environment was built using Python 3.11. For text embedding, we utilized Hugging Face Sentence Transformers as the embedding model framework.

\end{document}